\documentclass[10pt,journal,compsoc]{IEEEtran}
\usepackage{graphicx}
\usepackage{wrapfig}
\usepackage[small]{subfigure}
\usepackage{float}
\usepackage{mathtools}
\usepackage{algorithm}
\usepackage{algorithmic}
\usepackage[normalem]{ulem}
\usepackage[tableposition=top]{caption}
\usepackage{url}
\usepackage{booktabs}

\usepackage{array}
\newcolumntype{?}[1]{!{\vrule width #1}}
\usepackage{color}
\usepackage{setspace}
\usepackage{diagbox}
\usepackage{enumitem}
\usepackage{caption}
\usepackage{makecell}
\usepackage{listings}
\usepackage{url}
\usepackage{makecell}
\usepackage{boondox-cal}
\usepackage{ragged2e}

\IEEEoverridecommandlockouts
\usepackage{cite}
\usepackage{dblfloatfix}
\usepackage{amsmath,amssymb,amsfonts}
\usepackage{graphicx}
\usepackage{textcomp}
\usepackage{xcolor}
\def\BibTeX{{\rm B\kern-.05em{\sc i\kern-.025em b}\kern-.08em
    T\kern-.1667em\lower.7ex\hbox{E}\kern-.125emX}}

\AtBeginDocument{%
  \providecommand\BibTeX{{%
    \normalfont B\kern-0.5em{\scshape i\kern-0.25em b}\kern-0.8em\TeX}}}

\newcommand{\systemname}{PURE}

\begin{document}

\title{PURE: Passive mUlti-peRson idEntification via Deep Footstep Separation and Recognition}

\author{~Chao~Cai,~\IEEEmembership{Student Member,~IEEE,}
        ~Ruinan~Jin,
        ~Peng~Wang,
        ~Liyuan~Ye, 
        ~Hongbo~Jiang,
        ~and~Jun~Luo,~\IEEEmembership{Senior Member,~IEEE}
\IEEEcompsocitemizethanks{
\IEEEcompsocthanksitem C. Cai and J. Luo is with the School of Computer Engineering, Nanyang Technological University, Singapore 639798. E-mail: \{chris.cai, junluo\}@ntu.edu.sg.
\IEEEcompsocthanksitem R. Jin is with the School of Information and Safety Engineering, Zhongnan University of Economics and Law, e-mail: jrnsneepy@gmail.com.
\IEEEcompsocthanksitem P. Wang and  L. Ye are with the School of Electronic Information and Communications,
Huazhong University of Science and Technology, China. E-mail: \{puhenglin,yeliyuan\}@hust.edu.cn.
\IEEEcompsocthanksitem H. Jiang (corresponding author) is with the College of Computer Science and Electronic Engineering, Hunan University. E-mail: hongbojiang2004@gmail.com.
}}

\markboth{IEEE Transactions on Mobile Computing,~Vol.~xx, No.~x, XX~2021}%
{Shell \MakeLowercase{\textit{et al.}}: Active Acoustic Sensing for ``Hearing'' Temperature under Severe Interference}

\IEEEtitleabstractindextext{
{\justify
\begin{abstract} 
Recently, \textit{passive behavioral biometrics} (e.g., gesture or footstep) have become promising complements to conventional user identification methods (e.g., face or fingerprint) under special situations, yet existing sensing technologies require lengthy measurement traces and cannot identify multiple users at the same time. To this end, we propose \systemname\ as a passive multi-person identification system leveraging deep learning enabled footstep separation and recognition. \systemname\ passively identifies a user by deciphering the unique ``footprints'' in its footstep. Different from existing gait-enabled recognition systems incurring a long sensing delay to acquire many footsteps, \systemname\ can recognize a person by as few as only one step, substantially cutting the identification latency. To make \systemname\ adaptive to walking pace variations, environmental dynamics, and even unseen targets, we apply an adversarial learning technique to improve its domain generalisability and identification accuracy. 
Finally, \systemname\ can defend itself against replay attack, enabled by the richness of footstep and spatial awareness. 
We implement a \systemname\ prototype using commodity hardware and evaluate it in typical indoor settings. Evaluation results demonstrate a cross-domain identification accuracy of over 90\%.
\end{abstract}
}
}

\maketitle

\section{Introduction}
User identification technology is an indispensable element for building smart indoor applications, such as elderly and child care, personalized custom service, and
surveillance in sensitive zones. As one example, a smart home recognizing its elderly host (also his/her walking direction) may predict his/her intentions and switch on home appliances (e.g., lights or TVs) accordingly. Taking supermarket as another example, identifying customer identities and hence retrieving the relevant shopping histories may enable a salesman (or simply an advertising system) to make more appropriate recommendations and thus more likely to cut a deal. Existing solutions for user identification, ranging from conventional computer vision~\cite{SphereFace,ArcFace,CenterLoss,FaceNet} and biometrics~\cite{BiometricSensing1,BiometricSensing2,BiometricSensing3} to the emerging WiFi sensing~\cite{WiFiSensing1,WiFiSensing2,WiWho,WiFiID}, have shown promising results for several important applications. 

However, the inherent limitations of existing solutions have prevented them from being widely applicable.
In particular, computer vision techniques require a good lighting condition and line-of-sight (LoS), while they may raise critical privacy concerns.
Biometrics often demand wearable instruments or users' active involvements, hence causing potential discomfort and inconvenience. WiFi sensing approaches leveraging gait profile or breathing pattern appear to be viable, but they often fail to work in practice due to the severe interference from WiFi's main function of communications and other co-spectrum devices. 
Recent developments on passive behavioral biometrics~\cite{Footstep2, Footstep3} (e.g., footstep-enabled identification~\cite{FootprintID,Footstep1})
can be promising alternatives to the aforementioned solutions as they incur no privacy issues and have less interference due to their limited sensing range. However, 
these approaches often require special sensing hardware~\cite{Footstep1,Footstep2,Footstep3} or multiple long-time footstep measurements~\cite{FootprintID}. Besides, they can only handle one user each time, rendering them less appealing for practical applications where multiple users may appear at the same time. 

In this paper, we revisit the footstep-enabled passive behavioral biometrics, and we propose \systemname\ (Passive mUlti-peRson idEntification) to achieve a lightweight user identification, exploiting both commodity sensing devices and up-to-date deep learning techniques. 
\systemname\ is built on a key observation that footsteps carry ``footprints'' unique to individuals and thus can be leveraged for effective user identification. These footprints can be passively captured by commodity acoustic sensing hardware, totally removing the need for active user involvements. \systemname\ aims to extract such footprint information from as few as a single step, thus enabling a much faster identification than existing gait-oriented systems. In addition, \systemname\ targets simultaneous multi-user identification, robustness against various background interference, as well as immunity to replay attack.

Implementing the above ideas into a practical system entails several technical challenges. 
First of all, background noise and interferences (especially voices) may overwhelm footsteps, 
significantly affecting the system performance. Both detecting and extracting footsteps under such strong interference can be a formidable task. 
Second, footstep variations, caused by factors such as different walking paces, can lead to distinctive features hence affect the identification performance. In addition,
footstep can carry environment-dependent information (e.g., floor material). Such domain conditions are largely irrelevant to individuals' unique footprints, and can thus seriously degrade the identification performance if not sufficiently removed. 
Last but not least, replay attack that records someone's footstep and cheats an identification system later by playing the recorded sounds could be a major hurdle for practical adoption in certain applications. 
\begin{figure*}[b]
    \centering
    \includegraphics[width=0.98\textwidth]{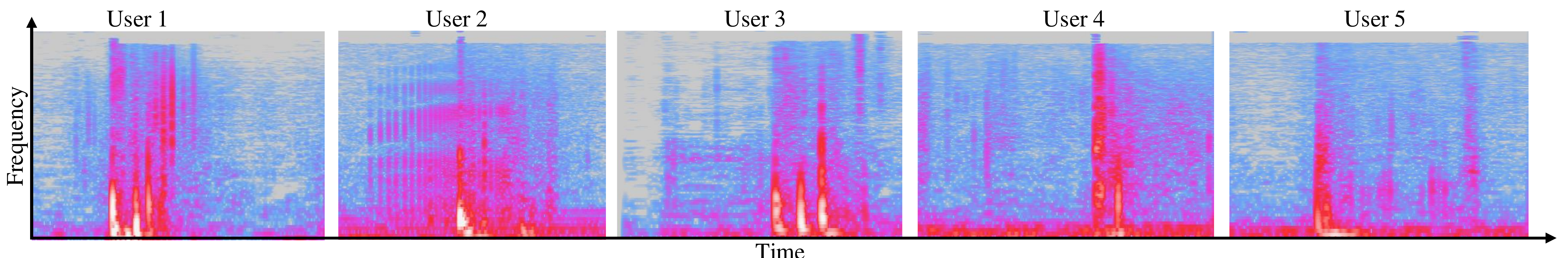}
    \caption{Mel-Frequency Cepstral  (MFC) of footsteps generated by five persons clearly show distinctive profiles.}
    \label{fig:mfcc of different users}
\end{figure*}

In~\systemname, we tackle these challenges via a series of delicate designs. In the presence of continuous voice signals, we explore the rhythmic patterns in the time-frequency (TF) representation to detect footsteps and employ a blind source separation algorithm with the \textit{a priori} knowledge to extract footsteps. 
To exclude feature variations caused by environment-dependent information and walking pace discrepancy for user identification, we train the user identifier (a predictor) via an adversarial domain adaptation scheme to improve its generalizability. 
Finally, we leverage the following fact to thwart replay attack: the replayed sounds exhibit static spatial characters (e.g., Angle-of-Arrival) and reveal inconsistency between walking speed and step frequency. 
To summarize, this paper makes the following contributions:
\begin{itemize}
    \item We propose \systemname, an acoustic passive multi-person identification system with little infrastructure cost.
    %
    %
    \item We innovatively employ an adversarial learning scheme to combat feature variations introduced by environment-independent information or heterogeneous walking paces, thus improving the system generalizability and identification performance.
    %
    \item We leverage the dynamic and smoothly changing spatial characters extracted from both structure-borne and air-borne footsteps to thwart the challenging replay attack.
    \item We implement \systemname\ prototype using commodity hardware and extensively evaluate its performance under various practical settings; the results demonstrate
    a cross-domain identification accuracy up to 90\%. 
\end{itemize}

The rest of this paper is organized as followed: Sec.~\ref{sec:background} discusses footstep basics. In Sec.~\ref{sec:system design}, we elaborate on the details of system design. Sec.~\ref{sec:performance evaluation} reports extensive performance evaluation results.
We present a literature review in Sec.~\ref{sec:related work}, and finally conclude the paper in Sec.~\ref{sec:conclusion}.

\section{Background and Motivation}
\label{sec:background}
We first explain the basic theories about footstep, then we further provide rationales for our later designs via a few preliminary measurement studies.

\subsection{Basic Acoustics of Footstep}
\label{ssec:basics}
When a foot touches the ground, it causes minor vibrations at the impact point and radiates energy via both the air and solid medium behind the surface. The acoustic pressure radiated from this impact point can be characterized by Rayleigh's surface integral~\cite{AcousticPhysics}:

\begin{small}
\begin{align}\label{eq:air-borne property}
 p\left(r, \theta, t\right) &= \frac{-2\rho}{\pi m d}\int_R \int_{\phi} R \sum _j \sum _k \left(-1\right)^{\frac{j-1}{2}}\sin \left[ j\pi \left( \frac{1}{2} + \frac{R}{a}\right) \right] \notag\\ 
    &\times \left[ \frac{\text{d}F}{\text{d}t} \left( t - d/c \right) * \cos \left( \omega_{jk} \left( t -d/c \right) \right) \right] \mathrm{d}R\mathrm{d}\phi,
\end{align}
\end{small}

\noindent where $(r, \theta)$ describes the position relative to the impact point with respect to the floor plane, $\rho$ is the mass density that characterizes the medium properties, $c$ is the speed of acoustic signals in a certain medium, $\omega$ is the frequency, $a$ is a constant, $d$ is the length of the leg, $F$ is the impact force being zero at all time except during the impact period, and $*$ denotes convolution. The acoustic pressure, generated by the impact event, mostly radiates through two common mediums. The one propagating through piston-like and non-dispersive air channel (we refer to as \textit{air-borne} hereafter) has a constant speed (e.g., a speed of 340 m/s at a temperature of 25$^\circ$C~\cite{AcousticPhysics}) and the corresponding waveform remains identical along the propagating path as far as there are no multipath reverberations. The other one traversing in solid medium, known as blending wave (we refer to as \textit{structure-borne} hereafter), exhibits a dispersive phenomenon where the propagation speed $c_f$ (ranging from 2000~\!m/s to 3000~\!m/s) of a specific signal component is a function of its frequency $f$:
\begin{equation}\label{eq:structure-borne property}
    {c_f} = \sqrt[4]{Eh{f^2} \left[12\rho (1 - v_p^2) \right]^{-1}},
\end{equation}
where $E, \rho, h$ are constants that characterize the property of a medium: $E$ quantifies the elastic property, $\rho$ is the mass density that indicates the stiffness, $h$ is the thickness, and $v_p$ is the phase velocity. Eqn.~\eqref{eq:structure-borne property} implies that when detecting structure-borne footsteps from different distances relative to the impact point, the corresponding waveform can be different due to the frequency-dependent propagation speeds.

\begin{figure}
    \centering
    \subfigure[Feature embeddings of different users under different environments.]{
    \label{fig:user and environment impacts}
    \includegraphics[width=0.42\columnwidth]{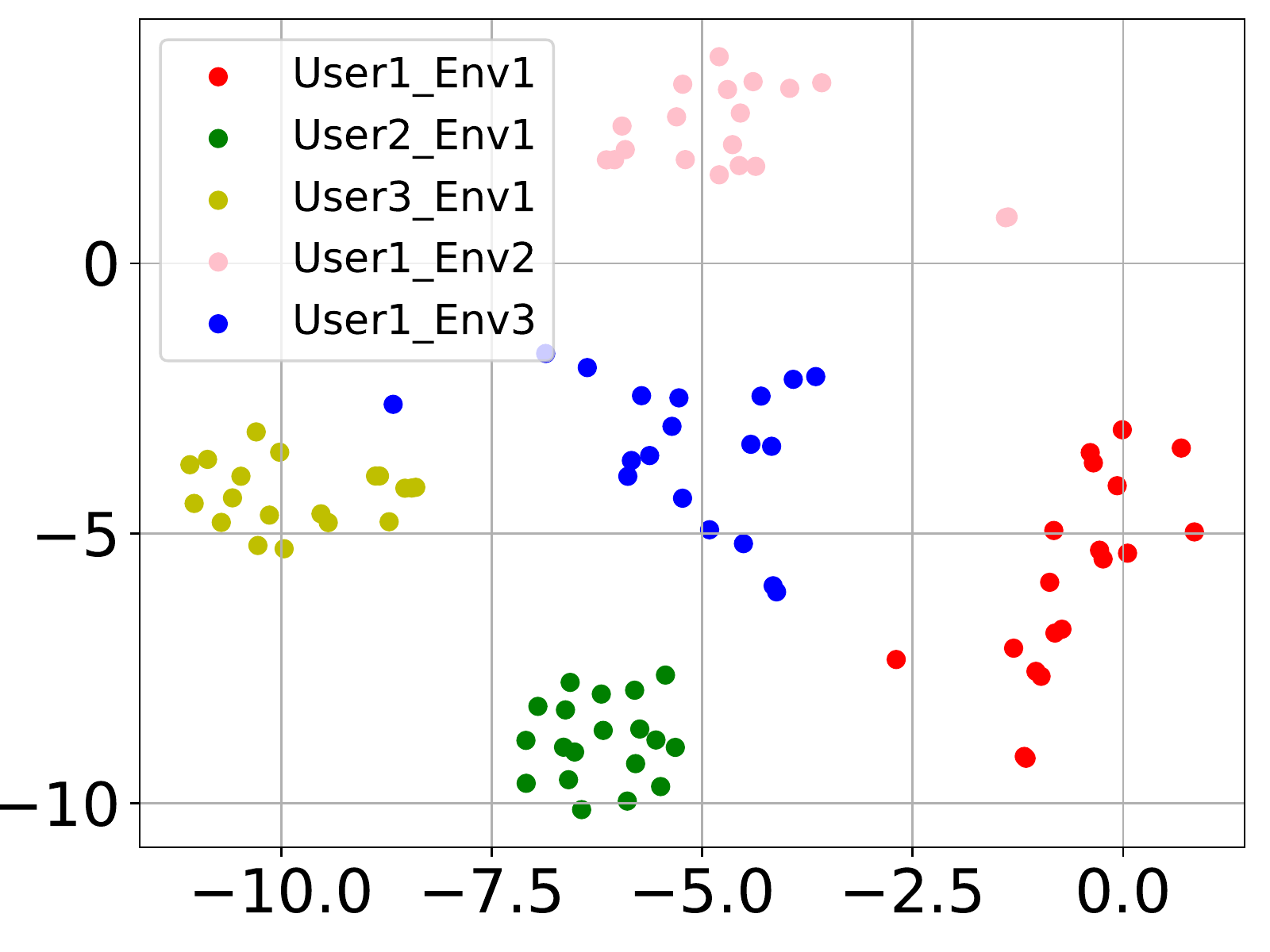}
    }
    \hspace{0.01\textwidth}
    \subfigure[Feature embeddings under walking pace variations. ]{
    \label{fig:walking pace variation}
    \includegraphics[width=0.42\columnwidth]{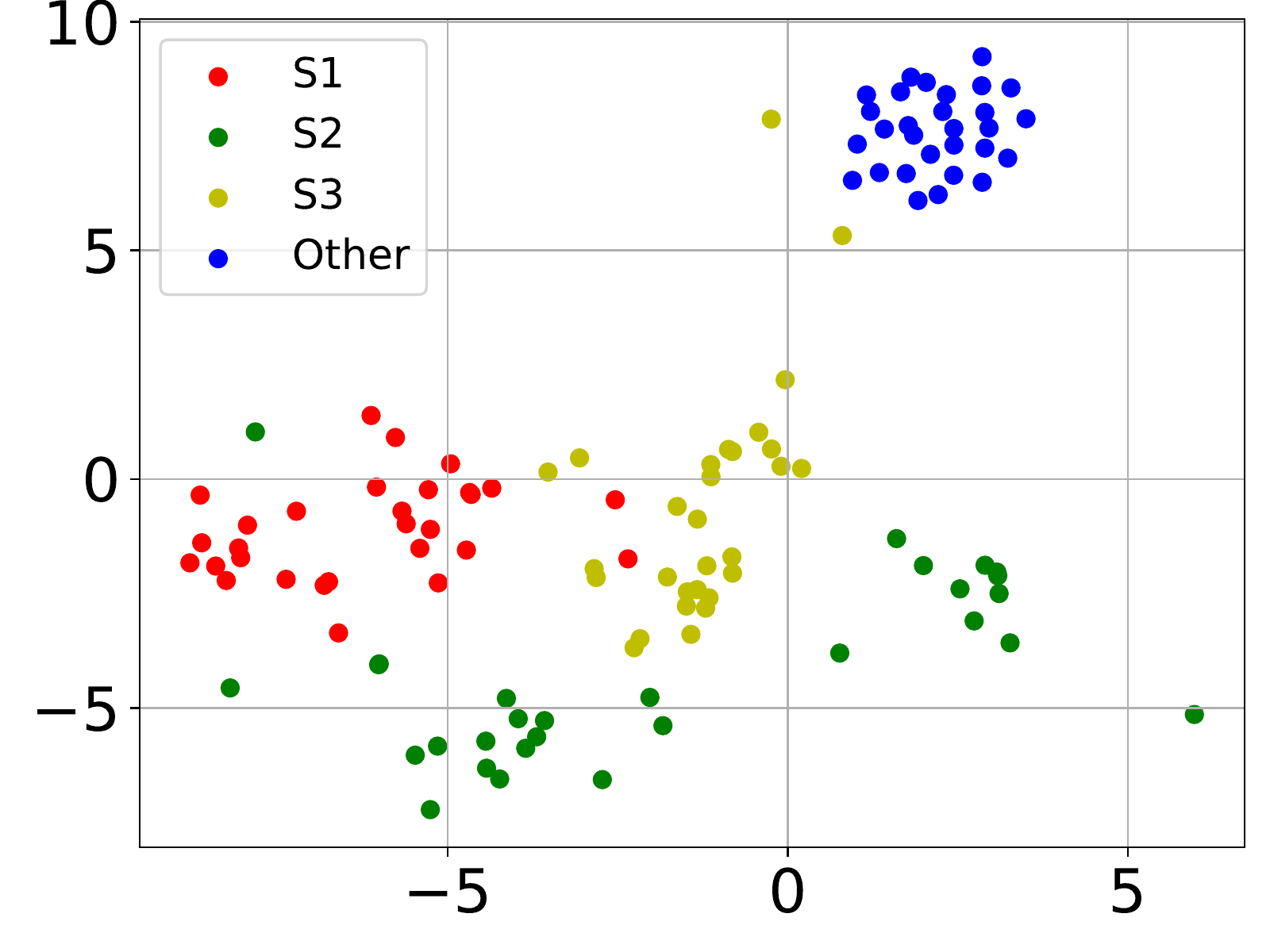}
    }
    \caption{
    MFC features embeddings (a) under different environments, and (b) under different walking speeds: s1 $=$ 0.2~\!m/s, s2 $=$ 0.5~\!m/s, s3 $=$ 1~\!m/s, ``other'' denotes other users at 0.2~\!m/s.}
    \label{fig:measurements}
\end{figure}

\subsection{Richness of Footstep Acoustic Profile}
%
%
Eqn.~\eqref{eq:air-borne property} and~\eqref{eq:structure-borne property} together show that footstep contains both air-borne and structure-borne signals and they involve a rich set of frequency components, carrying unique identity information. For example, both a person's weight (impact force) and walking style (duration of impact) are closely related to the generated acoustic pressures.
To verify the above intuition, we record footsteps from five persons under the same circumstance and inspect the corresponding Mel-Frequency Cepstrals (MFC)~\cite{MFCC}. The MFC features depicted in Fig.~\ref{fig:mfcc of different users} clearly show distinctive visual clues and thus imply the possibility of using footsteps for person identification. We then carry out measurements under different environments and Fig.~\ref{fig:user and environment impacts} plots the embedded features. One may observe that the low-dimensional features among different users show clear boundaries, demonstrating the possibility of effective identification using footsteps. We also conduct a simple user study by playing several recording of footsteps produced by different persons; all audiences participating in our user study indeed claim that they can tell perceivable differences among these recordings. The distinctiveness of acoustic features in footsteps from different persons lays the foundation for our footstep enabled user identification system. However, the footstep features of the same person can be affected by 
\textit{domain conditions} such as varying walking paces as demonstrated by Fig.~\ref{fig:walking pace variation}. These feature distinctions, introduced by domain conditions (including, e.g.,  environment heterogeneity and walking pace variations), can be particularly detrimental to the identification accuracy and thus should be excluded.

\subsection{Differences between Structure- and Air-borne Footsteps}
\begin{figure}[t]
    \centering
    \subfigure[Identification accuracy comparison.]{
    \label{fig:identification comparison}
    \includegraphics[width=0.42\columnwidth]{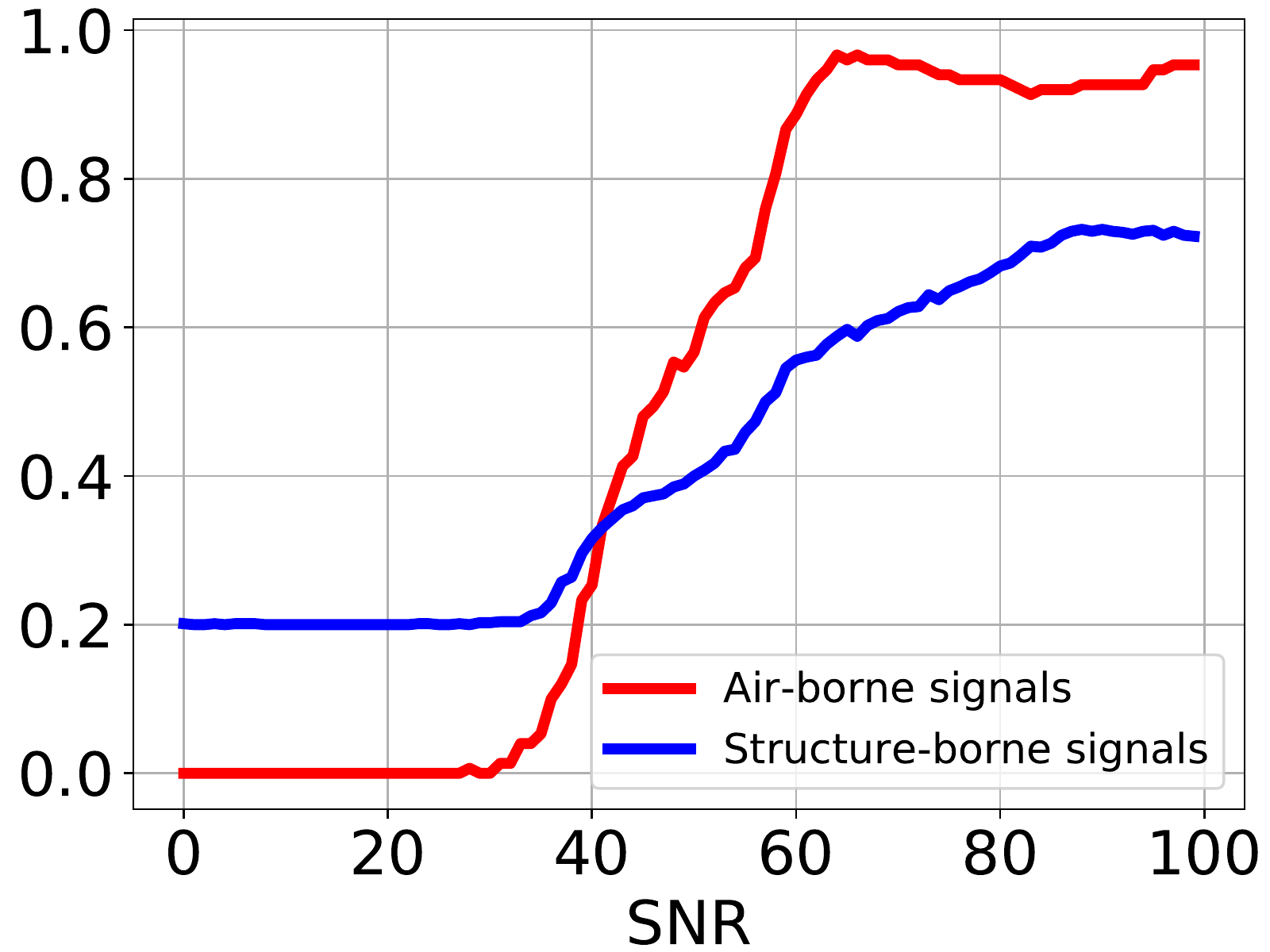}
    }
    \hspace{0.001\textwidth}
    \subfigure[Feature embeddings from different locations.]{
    \label{fig:structure-borne measurements}
    \includegraphics[width=0.42\columnwidth]{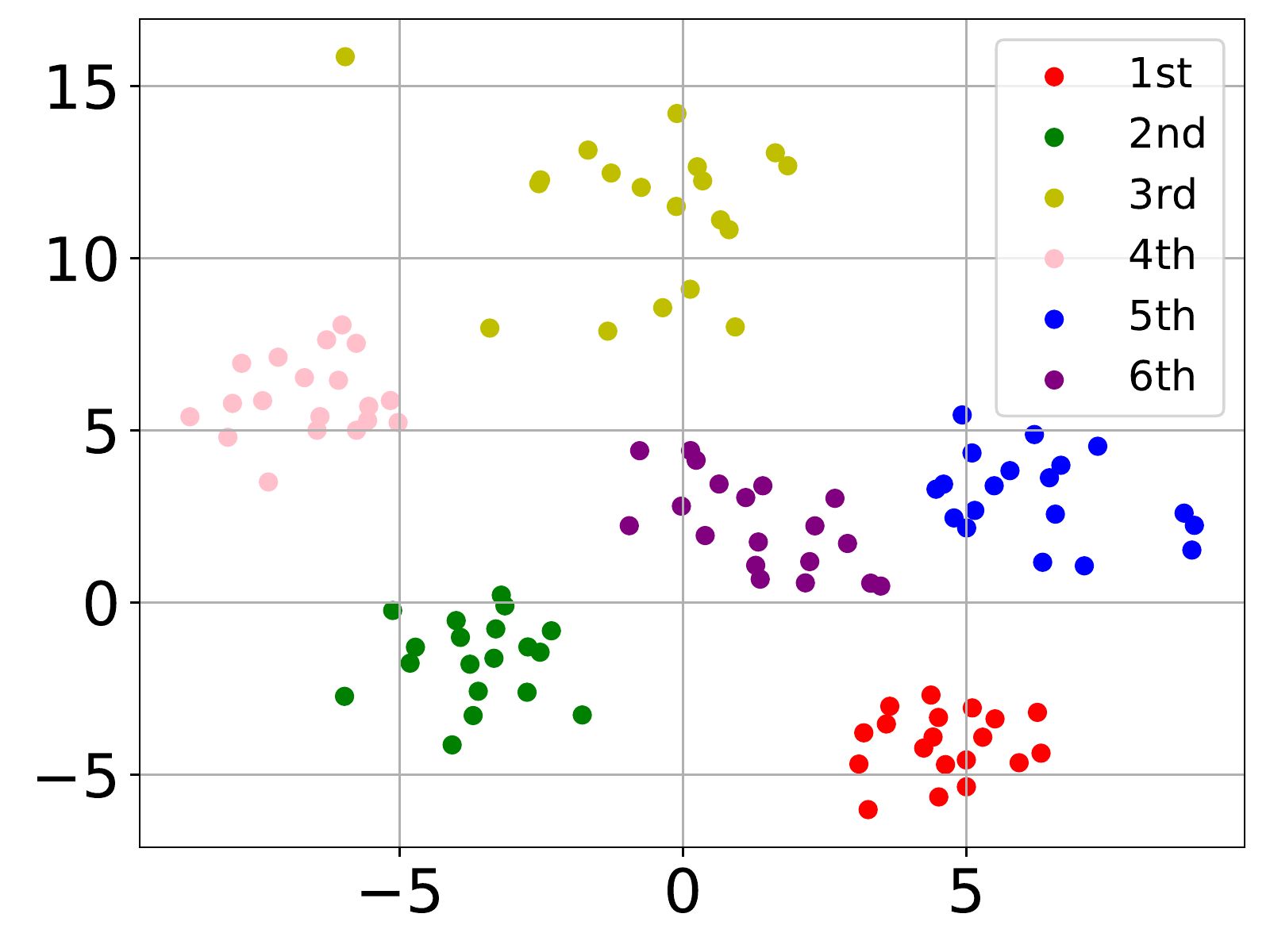}
    }
    \label{fig:structure-borne measurements identification and ranging}
    \caption{Examining the properties of structure-borne and air-borne components of footsteps. (a) GMM applied on air- and structure-borne signals for identification. (b) Structure-borne component carries distance information.}
\end{figure}

The structure-borne and air-borne components of footsteps have a sharp difference in their propagation speeds, as already discussed in Sec.~\ref{ssec:basics};
this offers us an opportunity to physically separate them. For instance, when the sampling rate is 192~\!kHz and we detect the footsteps at a distance of 2~\!m, a clean set of structure-borne signals (due to propagation difference) lasts for $\frac{1}{340} - \frac{1}{3000} \approx 5.4$~\!ms, an equivalent of 1000 samples. 
%
\begin{figure}[htp]
 	\centering
 	\includegraphics[width=.58\columnwidth]{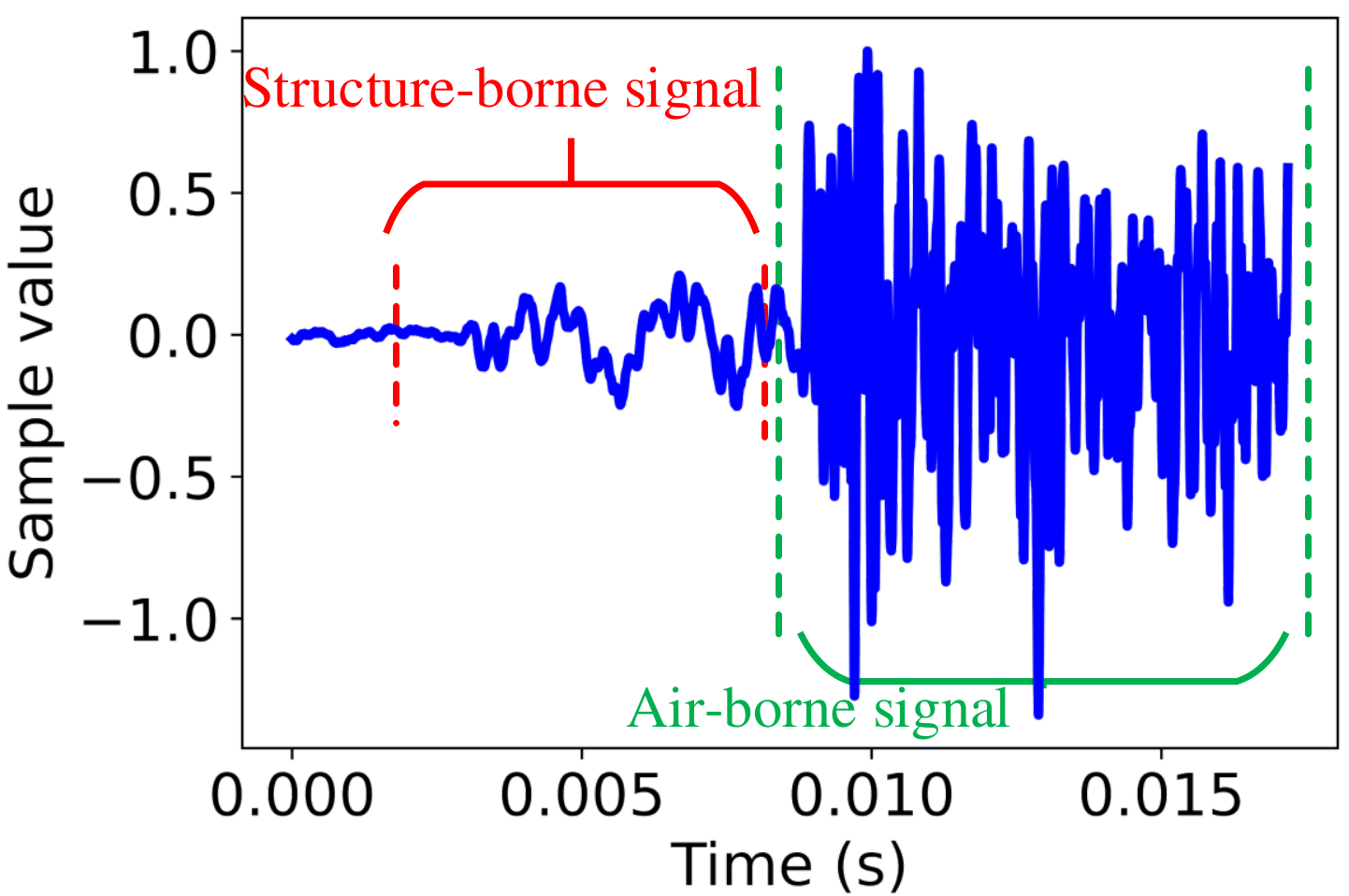}
 	\caption{A footstep waveform contains both structure- and air-borne components that are temporally separated.}\label{fig:footstep waveform}
\end{figure}
%
As clear shown by the measurements illustrated in Fig.~\ref{fig:footstep waveform}, the structure-borne component indeed arrives ahead of the air-borne one.
Separating these two components allows us to exam their respective natures. As the air-borne component appears to have more complicated waveforms, we conjecture that it may be more suitable for the identification purpose than its structure-borne counterpart.
To verify the above hypothesis, we have conducted measurements on the five persons in several environments using commodity microphones.  
We then utilize a Gaussian Mixture Model (GMM)~\cite{GMM} to identify those persons based on either structure-borne or air-borne components. The identification accuracy comparison under different Signal-to-Noise-Ratio (SNR) is shown in Fig.~\ref{fig:identification comparison}, which confirms the advantage of adopting the air-borne component for achieving a higher identification accuracy. Also, the experiments reveal that using traditional GMM method for identification is practically not feasible since the required SNR is over 60~\!dB. 

\subsection{The Edge of Structure-borne Footstep}
Although structure-borne footstep is less representative than its air-borne counterpart for identifying persons, it has a unique property, namely acoustic dispersion, as shown by Eqn.~\eqref{eq:structure-borne property}. The dispersion phenomenon indicates that the structure-borne waveform is modulated by distance, suggesting the possibility of using it for ranging. Our measurements in Fig.~\ref{fig:structure-borne measurements}, showcasing clear boundaries between feature embeddings of structure-borne footsteps from different locations, further demonstrate the feasibility of using structure-borne footsteps for ranging. This range information, together with the angle of signal arrival, may give us an edge over attackers who replay footstep recordings, because the smoothly changing spatial characters of human footsteps cannot be fully imitated.

\begin{figure}[t]
    \centering
    \includegraphics[width=0.92\columnwidth]{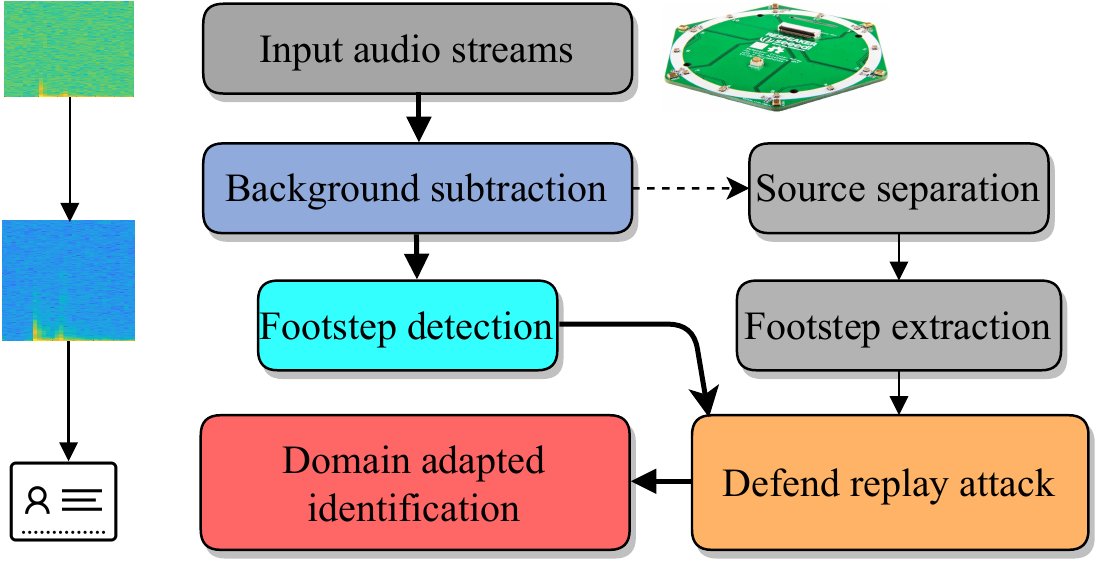}
    \caption{The system architecture of PURE.}
    \label{fig:system structure}
\end{figure}

\section{System Design}
\label{sec:system design}
PURE consists of a pipeline of signal processing and deep learning modules as shown in Fig.~\ref{fig:system structure}. It employs a microphone array~\cite{MicrophoneArray} to capture raw audio streams and then denoise them via a background spectral subtraction, removing most stationary noises. Then footstep detection is performed by inspecting the energy change, as well as by adopting an audio classifier. During the above process elaborated in Sec.~\ref{ssec:background suppression}, 
if there exist continuous interferences such as voices, a source separation algorithm kicks in to decouple footsteps from the sound mixture, as detailed in Sec.~\ref{ssec:footstep extraction under continous interference}. 
Finally, PURE applies a neural network to identify the user behind each footstep (Sec.~\ref{ssec:domain adapted identification}), and it also extracts spatial information to counter replay attack (Sec.~\ref{ssec:prevementing replay attack}).

\subsection{Background Noise Suppression and Footstep Detection}
\label{ssec:background suppression}

Indoor places often contain common color noises~\cite{BackgroundSubtraction} that can affect the SNR of captured footsteps and thus should be removed. From the measurements in Fig.~\ref{fig:measurements}, we can see that the spectra of footsteps almost spread over the entire available bandwidth. Therefore, using common bandwidth oriented filter to trim the out-of-band noise is infeasible. To this end, we utilize a multi-band spectral subtraction method~\cite{BackgroundSubtraction} to suppress background noise and obtain about 3~\!dB gain in SNR. 

\begin{figure}[b]
 	\centering
    \includegraphics[width=0.6\columnwidth]{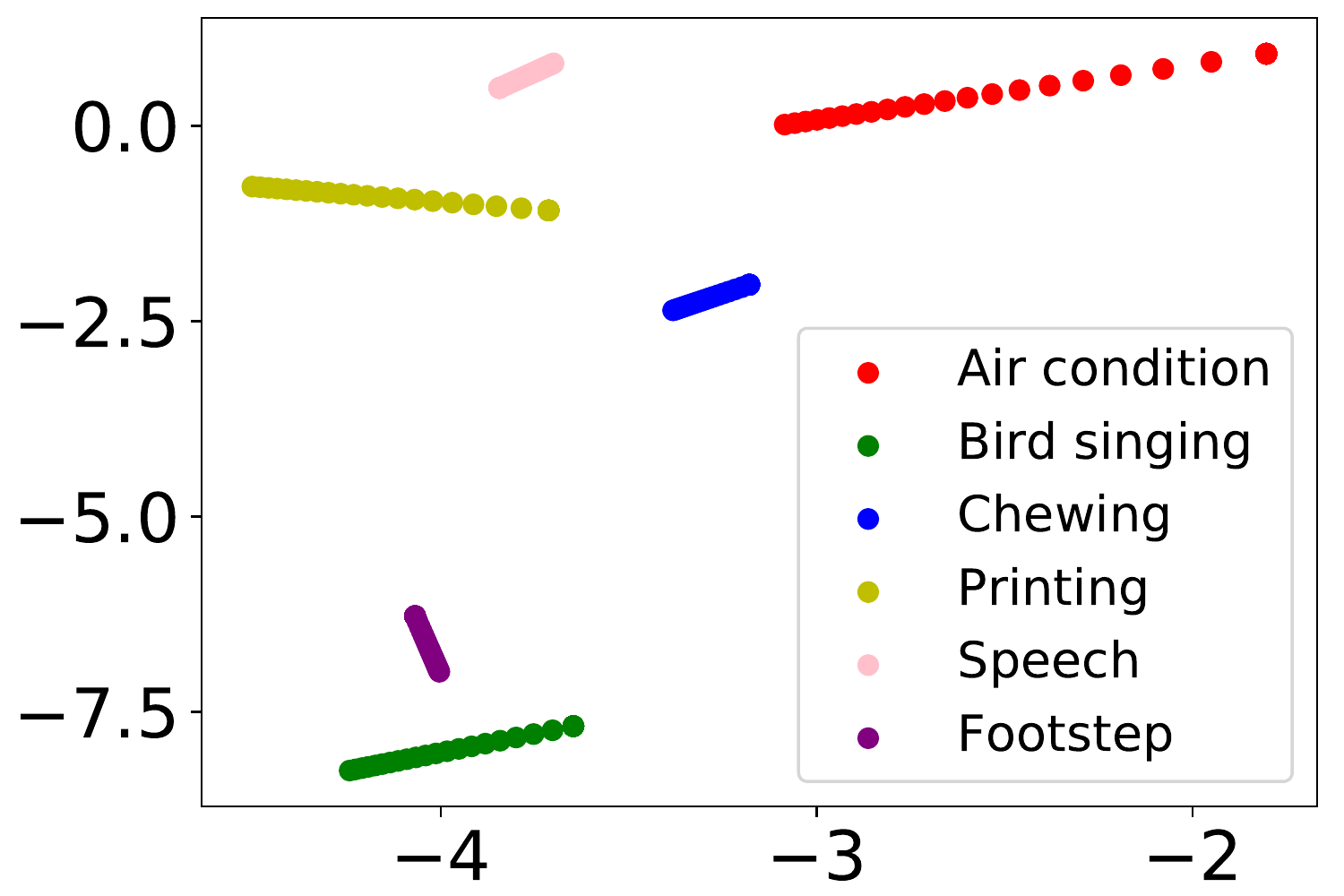}
    \caption{Low-dimensional outputs of GMM classifier show clear boundaries between different acoustic noises.}
    \label{fig:gmm clusters}
\end{figure}
%
After that, we apply abrupt energy detection, characterized by Root Mean Square (RMS)~\cite{RMS}, for candidate footstep detection. The RMS of a sequence $\mathbf{x} = \{ x_1, x_2, \cdots, x_L\}$ is defined by $E_\mathrm{RMS}(\mathbf{x}) = \sum _{i=1}^{L} \sqrt {\frac{x^2_1 + x^2_2 + ... + x^2_L}{L}}$. If the detected energy $E_{RMS}$ is above a certain threshold, a footstep  may be captured. Since many transient noises can also exhibit high energy, a simple energy-based detection method would not be sufficient. 
To this end, we further utilize a Gaussian Mixture Model (GMM)~\cite{GMM} based audio classifier to recognize footsteps. We train this GMM classifier against common background noises and it almost achieves an oracle performance: as shown in Fig.~\ref{fig:gmm clusters}, different acoustic signals can be clearly identified.


\subsection{Footstep Extraction under Continuous Interference }
\label{ssec:footstep extraction under continous interference}
To extract footstep overwhelmed by continuous strong interference such as voice signals, we further utilize a Flexible Audio Source Separation Toolbox (FASST)~\cite{FASST}. This procedure is often suspended for the sake of efficiency; it is only invoked when footsteps are heavily interfered. To leverage FASST, we use the rhythmic pattern of footsteps in frequency domain to detect the ongoing walking paces. More specifically, we utilize auto-correlation of STFT magnitude to detect the presence of footsteps. The reason for not directly using auto-correlation in time-domain is that the repetitive features in time-domain waveform are likely to be under noise floor due to their relatively low volume.

\begin{figure}[h]
    \centering
    \subfigure[Time domain correlation.]{
    \label{fig:waveform correlation content}
    \includegraphics[width=0.44\columnwidth]{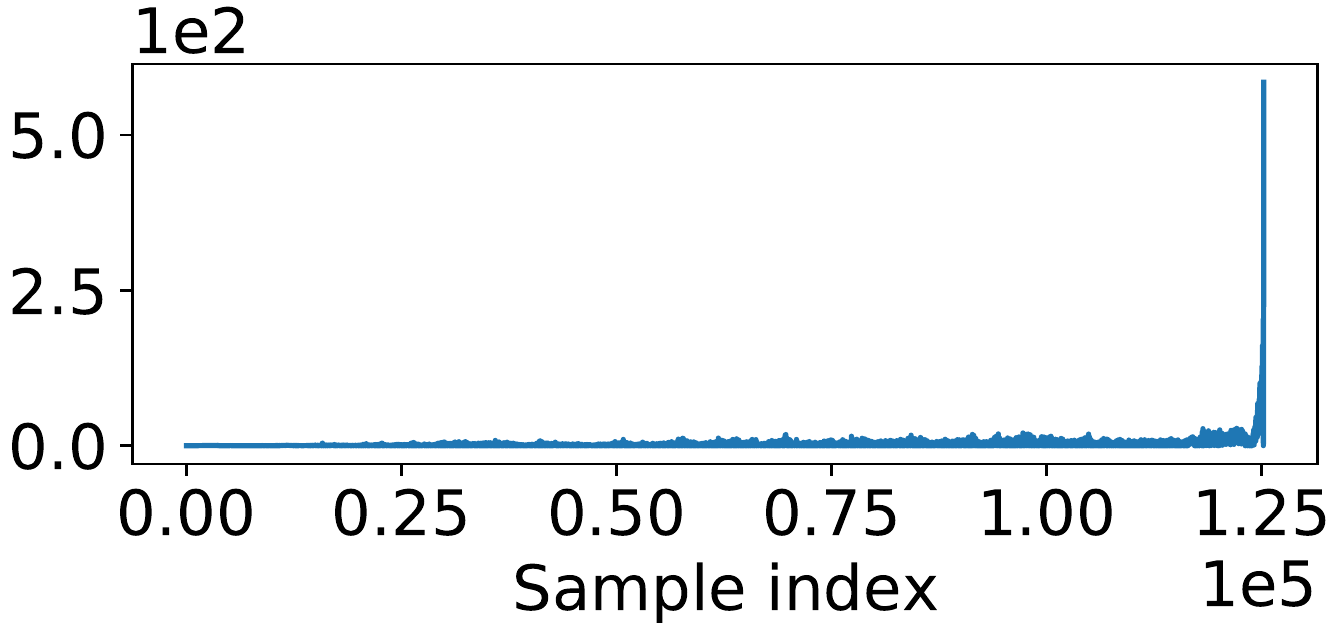}
    }
    \hspace{0.001\textwidth}
    \subfigure[Spectrum domain correlation.]{
    \label{fig:spectrogram correlation demonstration}
    \includegraphics[width=0.45\columnwidth]{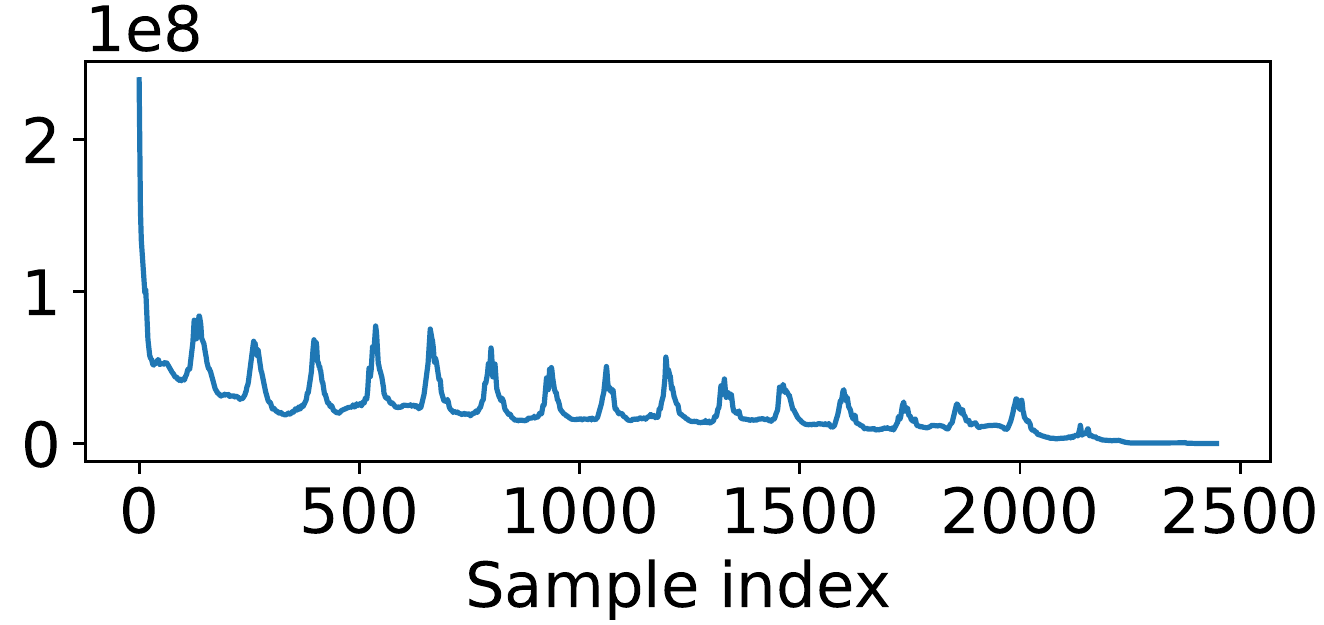}
    }
    \caption{Time domain correlation of a mixture of footsteps and voice fails to reveal rhythmic features (a), while in frequency domain, such features can be clearly observed (b). }
    \label{fig:correlation demonstration}
\end{figure}

Suppose $\mathbf{V}, \mathbf{V} \in \mathbb{R}^{P \times Q}$ denotes the STFT power spectrum where $P, Q$ are the dimensions of time and frequency, respectively, we first calculate the auto-correlation of $\mathbf{V}$ along the time frames to obtain $\mathbf{B}$, the operation of which can enhance the rhythmic pattern of footsteps: 
\begin{equation}\label{eq:detection in frequency domain}
    \mathbf{B}(i, j) = \frac{1}{P-j+1} \sum _{k=1}^{P-j+1} \mathbf{V}(i, k) \mathbf{V}(i, k+j-1). 
\end{equation}
We then take the average of $\mathbf{B}$ over the frequency dimension and normalize the result with its first term, $b(j) = \frac{1}{Q}\sum _{i=1}^{Q}\mathbf{B}(i, j), b(j) = \frac{b(j)}{b(1)}$. The walking rhythmic features can then be inspected in $\mathbf{b}=[b(j)]$, which we denote as the Averaged Spectrogram Auto-Correlation Coefficient (ASACC). We further check whether the periodicity exhibited in $\mathbf{b}$ lies in a reasonable range as the frequency of normal walking pace is usually within $[0.8, 2]$Hz~\cite{Zee}. Ideally, we can utilize $\mathbf{b}$ to obtain a soft binary mask for $\mathbf{V}$ and then extract the footsteps in the same vein as the proposal in~\cite{REPET}. However, as explain by~\cite{FASST}, the mask-based approaches, even with ground truth labels, are inferior to the local instantaneous Gaussian mixture models in source separation. Therefore, we still resort to FASST for better separation performance. Our measurements in Fig.~\ref{fig:correlation demonstration} further verify the above intuition as correlation in time domain shown in Fig.~\ref{fig:waveform correlation content} exhibit no rhythmic features while the rhythm is evident in frequency domain as shown in Fig.~\ref{fig:spectrogram correlation demonstration}. To further remove the residual noises in the separated signals, we apply a trained time-domain denoising network in~\cite{DEMUS} for footstep enhancement.

\begin{figure}[t]
    \centering
    \includegraphics[width=0.93\columnwidth]{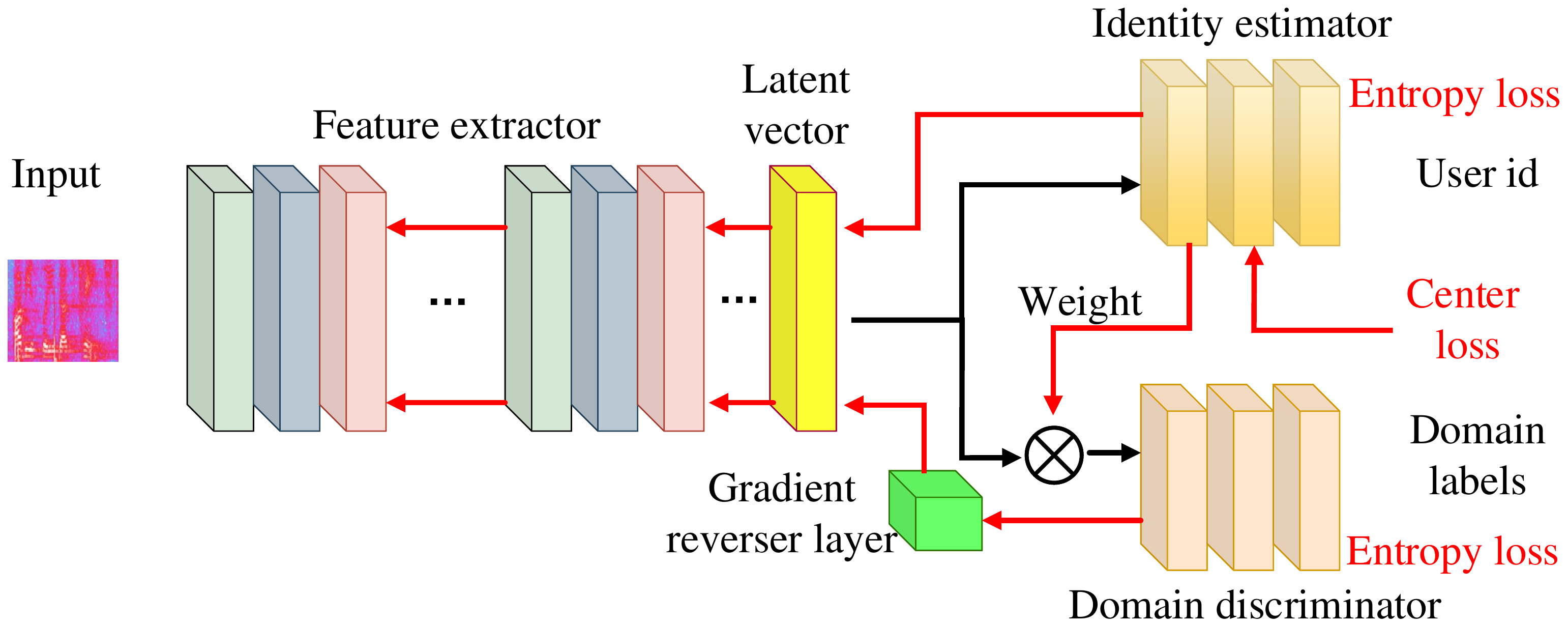}
    \caption{Identification network (ID-Net) based on domain adversarial adaptation.}
    \label{fig:DAT}
\end{figure}
\subsection{Domain Adapted Identification}
\label{ssec:domain adapted identification}

As we have mentioned in Sec.~\ref{ssec:basics}, the domain conditions, including walking speed variations and environment dynamics, can be particularly detrimental to our identification system. To preclude these domain information hence improve identification accuracy, we adopt domain adversarial adaptation. 
%
%
We formulate the identification problem as a \textit{classification problem}: given a specific footstep $\mathbf{x} \in {X}$, we aim to learn a maximum likelihood estimator $\mathcal{G}: {X} \rightarrow {I}$, where $X$ and $I$ represent the spaces of input (STFT of footstep waveform) and user identity, respectively.
In reality, an footstep $\mathbf{x}$ is sampled from a joint distribution $P(\mathbf{x}, u, d, v, e)$, where $u \in I$ that characterizes footstep diversity, $d \in D$ denotes distance information, $v \in V$ represents walking pace, and $e \in E$ describe specific properties incurred by environment dynamics. We denote $d, v, e$ potentially harmful to our identification purpose as \textit{domain conditions}. Apparently, only features characterizing the joint distribution $P(\mathbf{x}, u)$ are desirable for our identification purpose but features induced by domain conditions $d, v, e$ should be eliminated. To this end, we use a deep neural network $G(\mathbf{x})$ to approximate $\mathcal{G}(\mathbf{x})$ and adopt an adversarial learning~\cite{DomainAdv} to train $G$ so as to preclude the impact from $d, v, e$.

Our ID-Net is made up of three parts, namely feature extractor, identity predictor, and domain discriminator, as shown in Fig.~\ref{fig:DAT}. The feature extractor $\mathbf{f} = G_f(\mathbf{x}, {\boldsymbol\theta}_f)$, parameterized by ${\boldsymbol\theta}_f$, compresses an STFF of footstep waveform $\mathbf{x}$ into $\mathbf{f} \in \mathbb{R}^Q$, a lower-dimensional feature vector. With $\mathbf{f}$, the identity predictor aims to recognize the user behind the footstep, 
while the goal of domain discriminator is to identify different domains. The input to the domain discriminator is $\mathbf{f}$ weighted by a latent vector extracted from the identity predictor. The goal of ID-Net is to extract a \textit{domain independent} representation $\mathbf{f}$ so as to i) achieve high-accuracy user identification and ii) deceive the domain discriminator to misidentify domains. It is this domain independent $\mathbf{f}$ (involving only identity-specific features) that allows a cross-domain generalization of ID-Net.

The identity predictor $\hat P_u = G_u(\mathbf{f}, {\boldsymbol\theta}_u)$ outputs a probability matrix, whose element $\hat p^{(i, j)}_{u}$ represents the probability of the $i$-th footstep belonging to the $j$-th user. The loss function for training the parameter ${\boldsymbol\theta}_u$ is categorical cross-entropy: 
%
\begin{equation}
    \mathcal{L}_u = -|I|^{-1}\textstyle{\sum _{i=1}^{|I|}\sum_{j=1}^{N_u}} \log \left(\hat p^{(i, j)}_{u} \right), 
\end{equation}
where $N_u$ denotes the number of users. To further force the identity predictor to learn features sufficiently discriminative and generalizable to identify unseen users, we introduce a center loss~\cite{CenterLoss} in training the identity predictor:
\begin{equation}
    \mathcal{L}_C = 0.5~\textstyle{\sum_{i = 1}^{|I|}} \left \| \mathbf{f}_i - \mathbf{c}_{z_i} \right \|^2_2,
\end{equation}
where $\mathbf{c}_{z_i} \in \mathbb{R}^Q$ is the center for the $z_i$-th class deep features, $\mathbf{f}_i \in \mathbb{R}^Q$ is the $i$-th deep feature, and the summation is performed over the input set $I$. We update $\mathbf{c}_{z_i}$ in a mini-batch (with size $m$) manner where the gradient of $\mathcal{L}_C$ with respect to $\mathbf{f}_i$ is calculated as $\mathbf{f}_i - \mathbf{c}_{z_i}$ with
$
      \mathbf{c}_j = \mathbf{c}_j + \frac{\sum^{|I|}_{i=1} \mathcal{I} (z_i = j)(\mathbf{c}_j - \mathbf{f}_i) }{1 + \sum ^{|I|}_{i=1}\mathcal{I} (z_i = j)} 
$,
$\mathcal{I}(z_i = j)$ is an \textit{indicator function} whose value is 1 if $z_i = j$; otherwise 0. We combine the center loss $\mathcal{L}_c$ with the categorical cross-entropy loss $\mathcal{L}_u$ to train the identity predictor as
$
      \mathcal{L}_\eta = \mathcal{L}_u + \lambda \mathcal{L}_c 
$,
where $\lambda$ is a scalar that balances the two losses. This loss function enables ID-Net's identity predictor to maximize inter-class margins and minimize intra-class distances, thereby improving its generalizability.  

The loss for training the domain discriminator $G_d(\mathbf{f}, {\boldsymbol\theta}_d)$ is also categorical cross-entropy: 
%

\begin{equation}
\mathcal{L}_\delta = -|I|^{-1} \textstyle{\sum _{i=1}^{|I|} \sum _{j=1}^{N_d}} \log(\hat {\delta} ^{(i,j)}),
\end{equation}
where $\hat\delta ^{(i,j)}$ is the output probability for the $i$-th footstep originating from the $j$-th domain and $N_d$ denotes the number of domains. 
%
The overall training process aims to minimize both $\mathcal{L}_\eta$ and $\mathcal{L}_{\delta}$ by tuning their respective network parameters $\mathbf{\theta}_u$ and $\mathbf{\theta}_d$. In the meantime, $\mathbf{\theta}_f$ is tuned to minimize $\mathcal{L}_\eta$ but maximize $\mathcal{L}_{\delta}$ (via gradient reversal). 
This procedure forces $\mathbf{f}$ to retain only user-specific properties but discard those induced by domains, allowing ID-Net to handle footstep samples taken from unseen domains.

\subsection{Thwarting Replay Attack via  Spatial Clues}
\label{ssec:prevementing replay attack}
As a user identification system, \systemname\ may be adopted in applications where security is a concern (e.g., assisting authentication and surveillance). Under these circumstances, \systemname\ needs to be resilient to potential attacks, among which replay attack is the most lethal one. Similar to all other acoustic-related identification methods (e.g., speaker recognition~\cite{podder2018speaker}), footsteps can be overheard (and recorded) by a microphone and then be replayed to attack the system in the sense of faking a certain user. Fortunately, footsteps contain dynamic and smoothly changing spatial clues induced by user movements. On the contrary, a recording clip of footsteps often exhibit only static spatial characters or may otherwise suggest abnormal trajectories, so it should be readily recognizable upon fully extracting the spatial clues contained in footsteps.  

To thwart replay attacks, we propose to leverage two pieces of spatial information, namely Time-of-Arrival (ToA) and Angle-of-Arrival (AoA), to filter out replayed footsteps. In particular, ToA is extracted from the structure-borne footstep and AoA is obtained from the air-borne counterpart. 
The key rationale is that footsteps from a lively walking person exhibit two traits: i) the moving speed suggested by ToA is bounded, as one should not move too fast indoors (it is also suspicious for a real person to move too fast in a security-concerned area) 
and ii) the moving trajectory implied by ToA should be naturally irregular, i.e., no person may move along a straight line.
On the contrary, an replayed footstep clip always violate at least one of these two criteria, as explained next, regardless of whether the attack is aware of our countermeasure or not. 

We formulate the defense against replay attack as a Hypothesis Test~\cite{Hypothesis}, in which we define significance thresholds to judge whether a footstep can be accepted as an authentic one or rejected as a replayed one. 
In our implementation, two thresholds are defined: i) we use Spearman coefficient~\cite{spearman} $\pi$ to quantify the correlation between walking speeds $\mathbf{v}$ and step frequency $\mathbf{f}$, as replayed footsteps often exhibit rather low correlation while those of authentic footsteps are high. ii) we exploit the maximum difference between any detected AoA $\gamma_\mathrm{diff}$ to check the motion states:
the AoAs should exhibit variance instead of being static. To be more specific, if $\pi \ge \bar\pi$ and $\gamma_\mathrm{diff} \ge \bar\gamma_\mathrm{diff}$ are both satisfied, we accept the footsteps; otherwise, we reject them, where $\bar\pi = 0.8$ and $\bar\gamma_\mathrm{diff} = 10^\circ$ are empirically set based on measurements.

We apply beamforming techniques~\cite{AudioBeamforming} to extract AoAs from footstep, but obtaining ToA, or equivalently range, is non-trivial. Intuitively, a range-sensitive acoustic fingerprinting strategy based on Sec.~\ref{ssec:basics} can be used, 
but such fingerprints can be interfered by domain conditions.
Therefore, we again resort to adversarial learning to preclude domain conditions with respect to ranging. 
%
Essentially, we train a neural network, R-Net, to infer range from a footstep $\mathbf{x}$.
This R-Net follows the same design as ID-Net except that the identity predictor is replaced by a range estimator (for regression) and the domain conditions are modified accordingly. We omit the training details given their similarity to those presented in Sec.~\ref{ssec:domain adapted identification}.

\begin{figure}[h]
    \centering
    \subfigure[Correlation between $\mathbf{v}$ and $\mathbf{f}$.]{
    \label{fig:replay attack}
    \includegraphics[width=0.46\columnwidth]{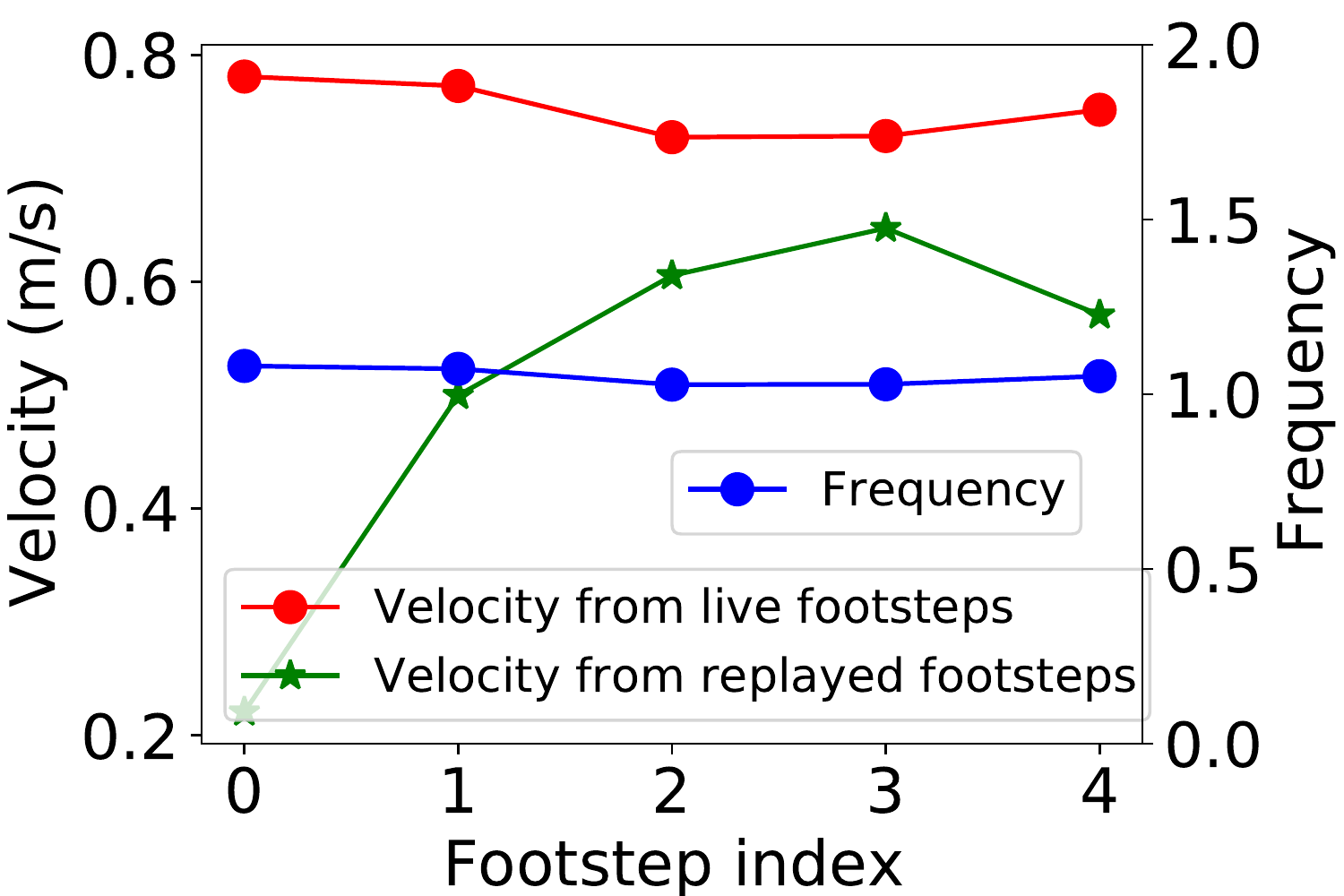}
    }
    \subfigure[AoA vs. range changes.]{
    \label{fig:spatial parameters}
    \includegraphics[width=0.46\columnwidth]{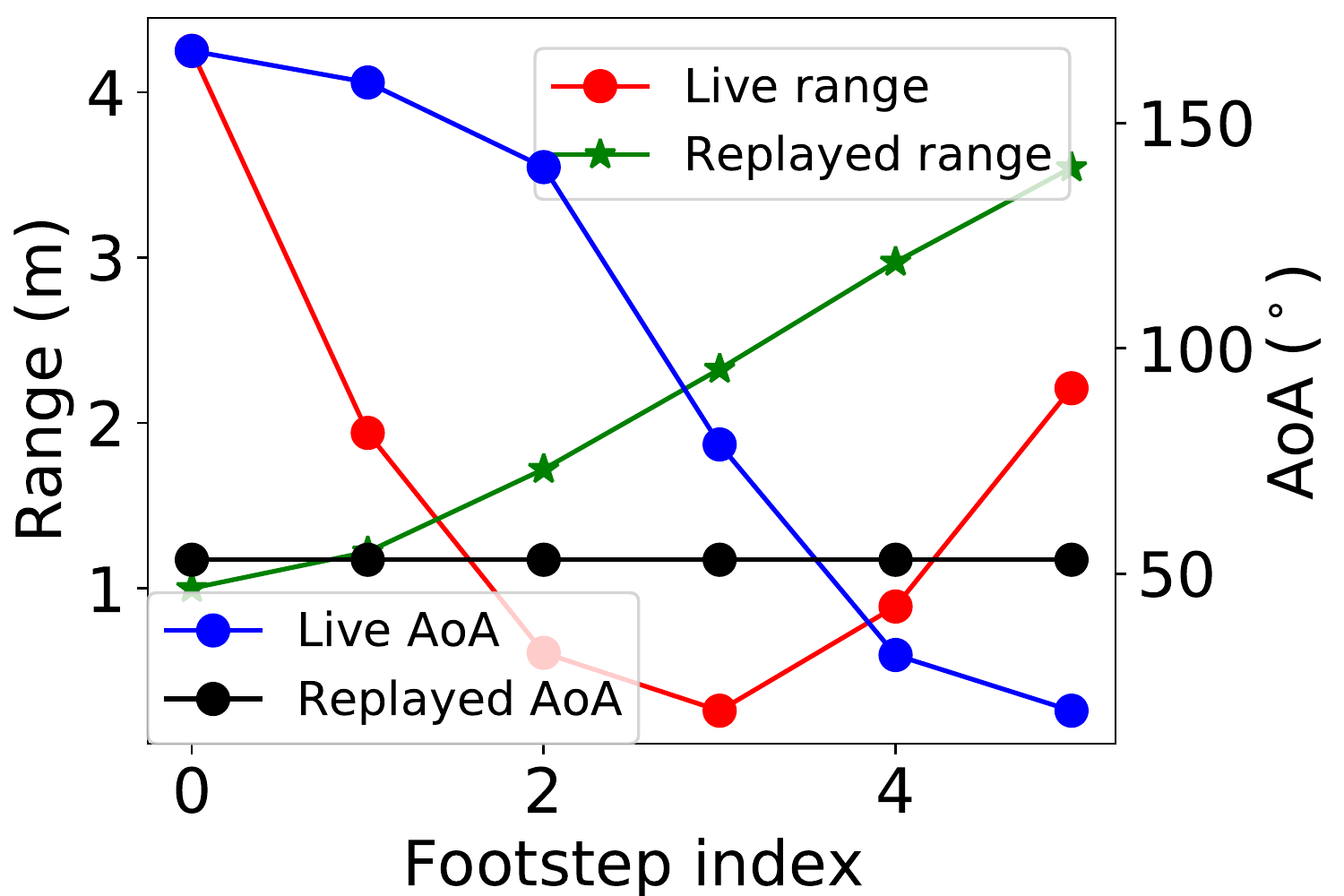}
    }
    \caption{Emulated scenario to compare (a) walking speed $\mathbf{v}$ vs. step frequency $\mathbf{f}$ (representing $\pi$) and (b) AoA vs. range changes (suggesting $\gamma_\mathrm{diff}$) between live and replayed footsteps.}
    \label{fig:replay attacks}
\end{figure}

%
We emulate a case to show how replayed footsteps exhibit abnormal properties and can thus be detected. 
We let a user (hence his/her footsteps) move from $[1, 0]$~\!m to $[1, 3.4]$~\!m, with a speed of 0.7~\!m/s (0.1~\!m/s variance) and a step frequency of 1~\!Hz (0.05~\!Hz variance). In the meantime, the microphone of \systemname\ is located at $[1.5, 2]$~\!m, while an attacker hides at the origin to record the footsteps. After a while, the attacker replays the recorded footsteps, trying to impersonate the legitimate user. As shown in Fig.~\ref{fig:replay attack}, $\mathbf{v}_\mathrm{replay}$ has a significantly lower correlation with $\mathbf{f}$, compare with that of $\mathbf{v}_\mathrm{live}$: $\pi_\mathrm{replay} = 0.32 < \bar{\pi} = 0.8 < \pi_\mathrm{live} = 0.87$.
Meanwhile, Fig.~\ref{fig:spatial parameters} shows a fixed $\gamma_\mathrm{diff} = 0$
for replayed footsteps, as opposed to the meaningful one for the live ones.
%

\section{Implementation and Performance Evaluation}
\label{sec:performance evaluation}
\begin{figure}[t]
    \centering
    \subfigure[A microphone array.]{
    \label{fig:microphone array}
    \includegraphics[width=0.44\columnwidth]{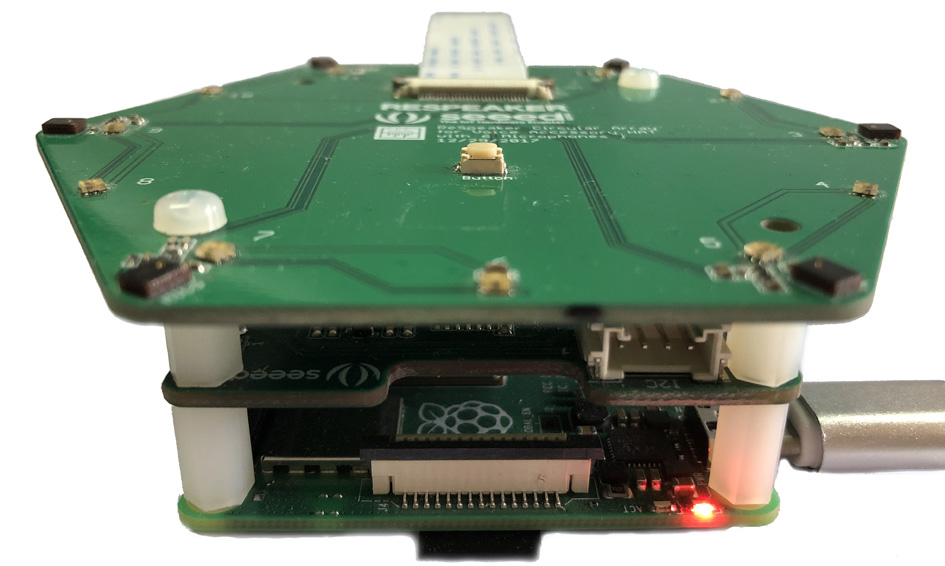}
    }
    \hspace{0.001\textwidth}
    \subfigure[Experiment setting.]{
    \label{fig:experiment setting}
    \includegraphics[width=0.43\columnwidth]{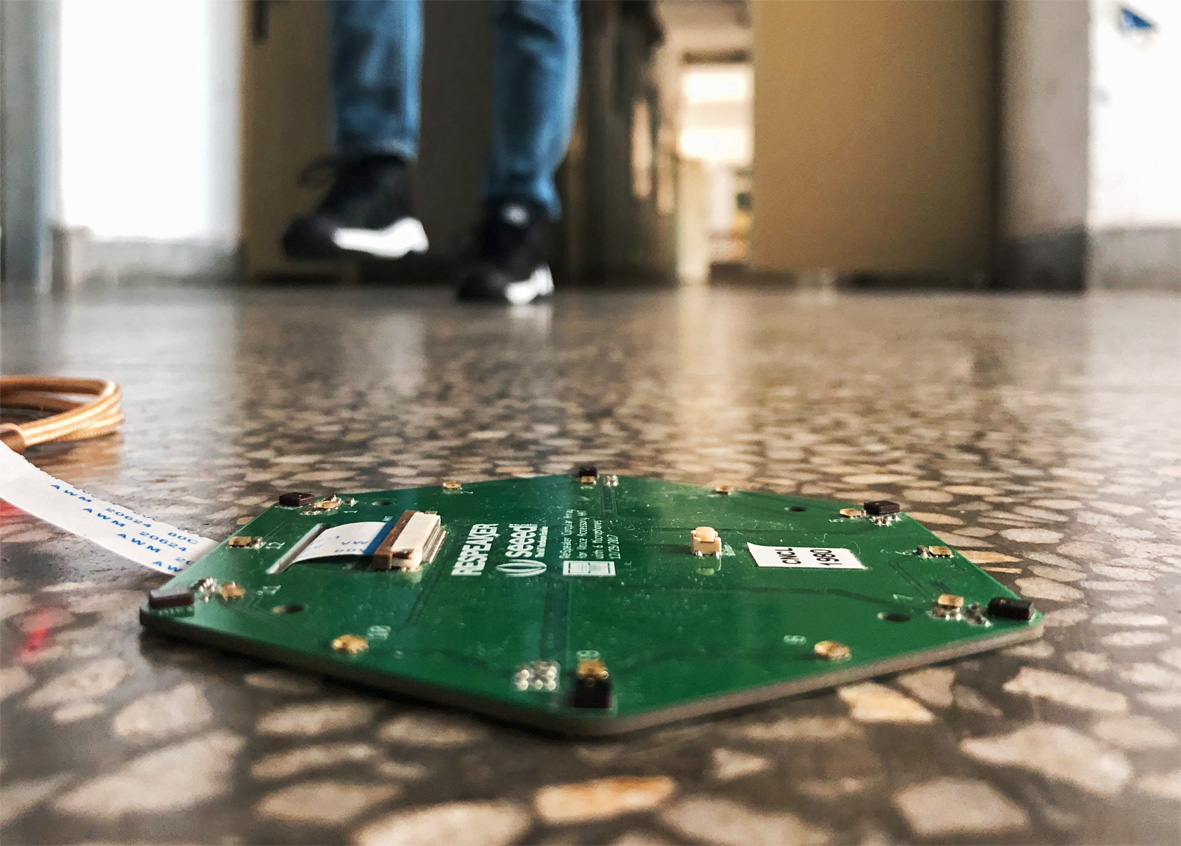}
    }
    \caption{Implementation with a microphone array (a) and corresponding experiment setting for evaluations (b).}
    \label{fig:microphone array and setting}
\end{figure}

\begin{figure*}[!t]
    \centering
    \subfigure[Clean footsteps.]{
    \label{fig:clean footstep autocorr}
    \includegraphics[width=0.6\columnwidth]{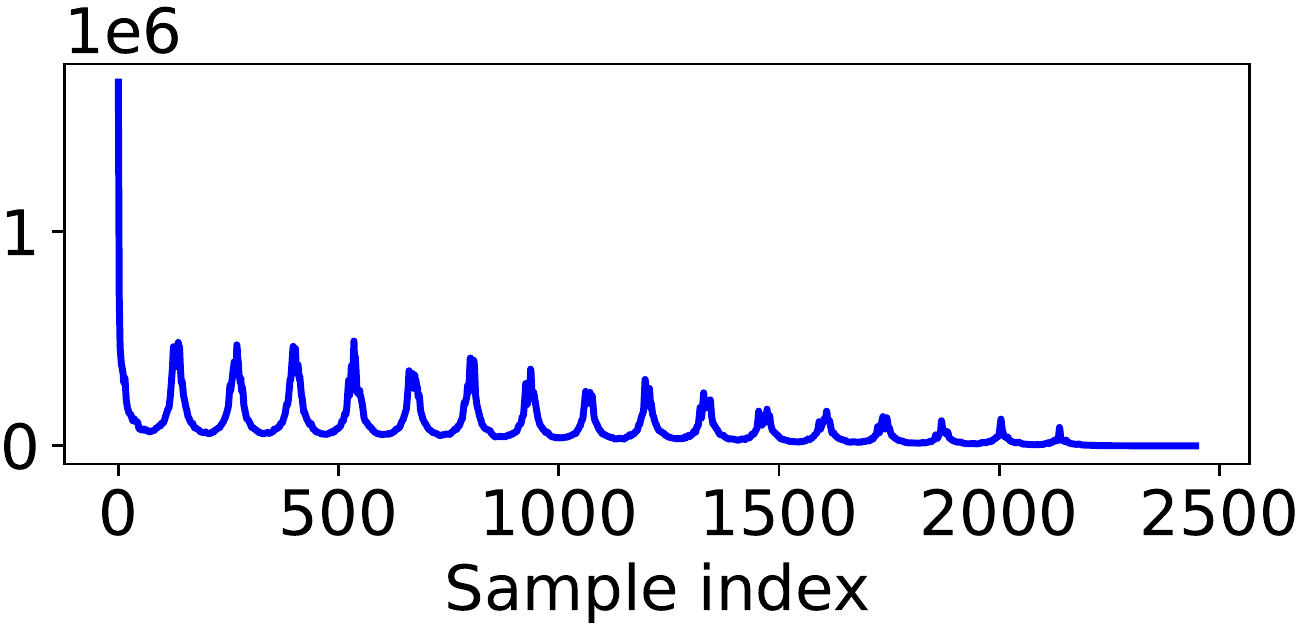}
    }
    \hspace{0.003\textwidth}
    \subfigure[Voice.]{
    \label{fig:voice autocorr}
    \includegraphics[width=0.6\columnwidth]{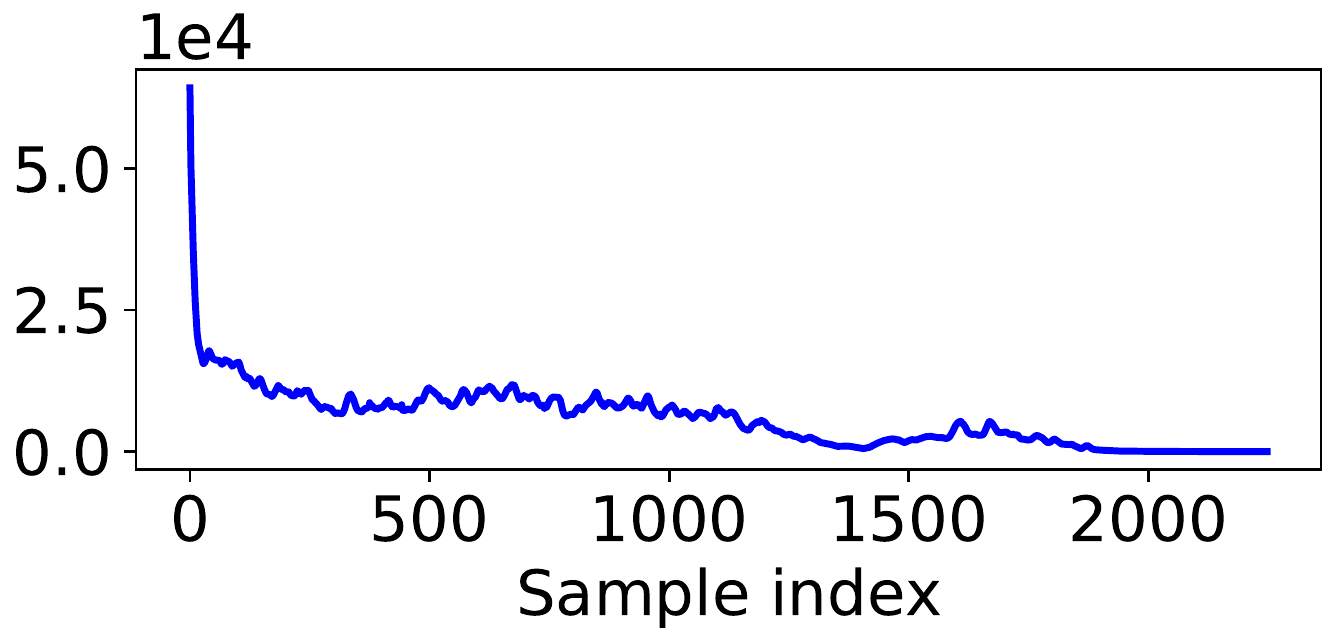}
    }
    \hspace{0.003\textwidth}
    \subfigure[Mixture.]{
    \label{fig:mixture autocorr}
    \includegraphics[width=0.6\columnwidth]{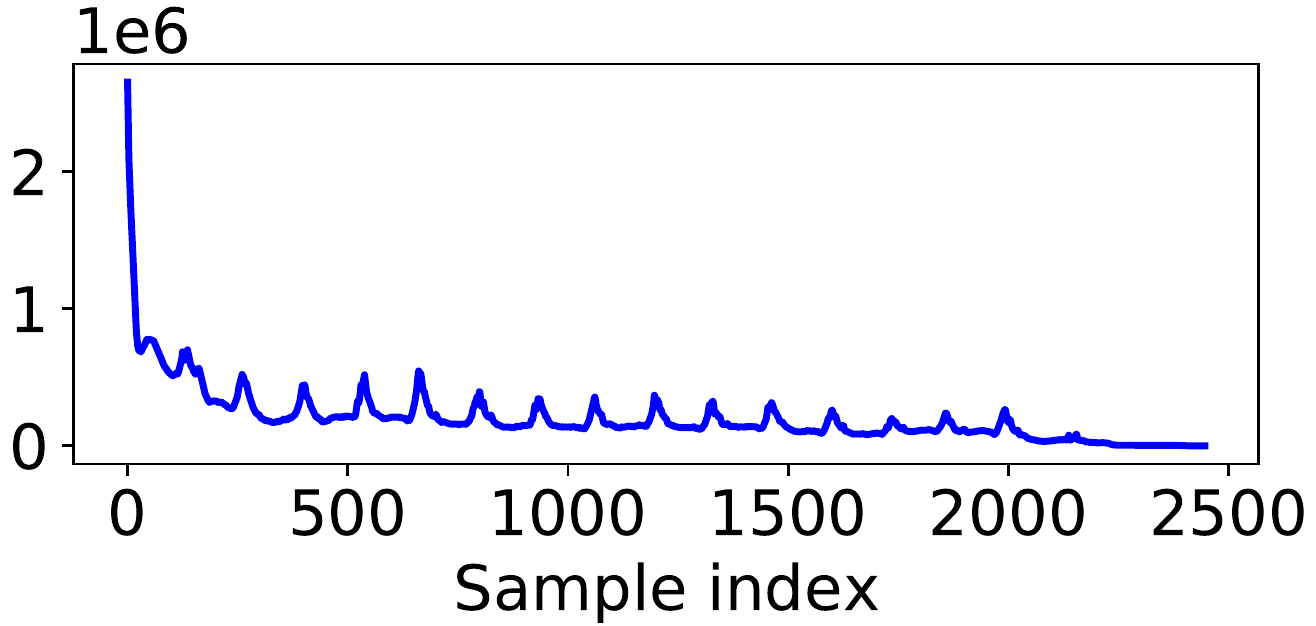}
    }
    \caption{ASACC calculated from (a) clean footsteps, (b) voice, and (c) a mixture of clean footsteps and voice.  }
    \label{fig:ASACC illustration}
\end{figure*}

\subsection{Implementation}
We implement a \systemname\ prototype using a circular array backed by a Raspberry Pi4, as shown in Fig.~\ref{fig:microphone array}. We configure the sampling rate as 192~\!kHz, the highest configuration on this platform, to capture as much structure-borne signals as possible due to their short duration. However, we downsample the air-borne footsteps to 16~\!kHz for identification purpose, in order to achieve computational efficiency. The signal processing modules (including footstep detection, GMM based classifier, and FASST audio source separation) are implemented using C++. The deep learning modules, namely the denoising network, ID-Net, and R-Net,
are implemented using Tensorflow~\cite{TF}. For the denoising network, we follow the routines in~\cite{DEMUS}, which allows \systemname\ to achieve a salient denoising performance and also a low runtime complexity.


ID-Net takes STFT magnitude of dimension $32\times16$ as its input, representing a footstep that lasts around $30$~\!ms.
The feature extractor has two convolution layers each followed by a batch normalization layer and ReLU activation layer. The first one has 32 filters with a $5\times 3$ kernel and the second one has 64 filters with a $3\times2$ kernel. After then, we flatten the output of the last convolution layer, add a dropout layer with drop probability of 0.65, and project it into a 16-dimensional feature vector. The identity predictor and domain discriminator both have only one fully connected layer with 16 neurons and adopt a sigmoid activation function. Their respective outputs depend on the number of users and domains involved in the training set.

For the R-Net, the input is a footstep waveform with 500 samples. The feature extractor in R-Net has a 1D convolution layer, followed by a pooling layer, and a fully-connected layer. The corresponding filter size is 64, 16, and 16, respectively. A dropout layer with a probability of 0.5 is inserted before the fully-connected layer. The range estimator, together with the domain discriminator has only one hidden layer, and both has a filter size of 16. To gather training data for R-Net, we first deploy a centimeter-level localization system using Decwave UWB-based sensors~\cite{Decwave}. We tie one sensor on a user's foot when he/she walks so as to obtain ground truth locations, i.e., ToA and AoA labels. 
Also, we attach an IMU sensor on the user's leg to help triggering the microphone array, so that it may correctly capture a footstep.

We synthesize a training data set for the denosing network with clean footsteps  from~\cite{Footstepsound1, Footstepsound2} and speech from TED talks~\cite{TEDtalks}. The noise is extracted from Diverse Environments Multichannel Acoustic Noise Database (DEMAND)~\cite{DEMAND}. We also collect footsteps under common floors (wood, stone) and circumstances (hall, indoor office, home appliance). We configure the signal SNR in a range of 5~\!dB, 10~\!dB, 20~\!dB, and 30~\!dB during training. We also utilize the footsteps from source separation to train the network in gaining the ability of minimizing residual interference signals.

\subsection{Performance Evaluation}
We present extensive experiments in this section. We start with evaluating the source separation algorithm, followed by the denoising network. Then we seriously verify the identification accuracy. Finally, we report performance of defending against the replay attack. The experimental statistics, unless otherwise noted, are all obtained by repeating the same experiment 1,000 times.


\subsubsection{Source Separation Performance}
As we mentioned in Sec.~\ref{ssec:footstep extraction under continous interference}, the source separation module is activated if we detect rhythmic features in STFT spectrogram. Therefore, before source separation, we first evaluate the performance of this detection algorithm. Recall that we utilize the ASACC $\mathbf{b}$, calculated by Eqn.~\eqref{eq:detection in frequency domain} for footstep detection, as the strongest energy of footsteps lies in the low frequency range, we thus only use the first three bins. As shown in Fig.~\ref{fig:ASACC illustration}, only footsteps that exhibit rhythmic features can generate periodic peaks in ASACC of Fig.~\ref{fig:clean footstep autocorr}, whereas voice signals hold no such properties in Fig.~\ref{fig:voice autocorr}. And when footsteps are mixed with voice, using ASACC can still identify this rhythmic feature as shown in Fig.~\ref{fig:mixture autocorr}. The detailed steps of this detection algorithm proceed as follows. 
After obtaining $\mathbf{b}$, we estimate the beating frequency $k$ in $\mathbf{b}$ via Discrete Fourier Transform. If the magnitude $m_{k}$ of bin $k$ in spectrum goes beyond the average of its subsequent 20 bins by a certain threshold (10 in our case), namely $m_{k} > {\frac{1}{20}\sum ^{k+20}_{i=k} m_i} + 10$, we accept that current audio signals contain footsteps; otherwise, no footstep is detected and source separation is deactivated. 
This detection method allows us to achieve a 100\% footstep detection even when the magnitude of voice is higher than footstep. 

To inspect the source separation performance, we first use clean footsteps blended with voice signals under different configurations as inputs and then check the quality of separated signals. Specifically, we synthesize mixtures of footstep and voice under different Source to Interference Ratio (SIR)~\cite{EvaluationMatric} as inputs and evaluate the performance using SIR and Source to Distortion Ratio (SDR)~\cite{EvaluationMatric}. These two evaluation matrices, namely SIR and SDR, are widely used to quantify source separation performance where $\mathrm{SIR} = 10\log\frac{\| \mathbf{s}_\mathrm{tgt} \|^2_2}{\| \mathbf{e}_\mathrm{itf} \|^2_2}$, $\mathrm{SDR} = 10\log\frac{\| \mathbf{s}_\mathrm{tgt} \|^2_2}{\| \mathbf{e}_\mathrm{itf} + \mathbf{e}_\mathrm{nie} + \mathbf{e}_\mathrm{atf}\|^2_2}$, with $\mathbf{s}_\mathrm{tgt}, \mathbf{e}_\mathrm{itf}, \mathbf{e}_\mathrm{nie}$, and $\mathbf{e}_\mathrm{atf}$ being respectively the target signal, interference signal, noise signal, and signal artifacts. The higher the value of these matrices, the better the target signal quality is. Both SIR and SDR in this experiment are calculated given footsteps as the primary signals, as opposed to common speech enhancement tasks where voices are the major concerns.

\begin{figure}[b]
    \centering
    \subfigure[SIR before separation vs SDR  after separation.]{
    \label{fig:sir-sdr}
    \includegraphics[width=0.43\columnwidth]{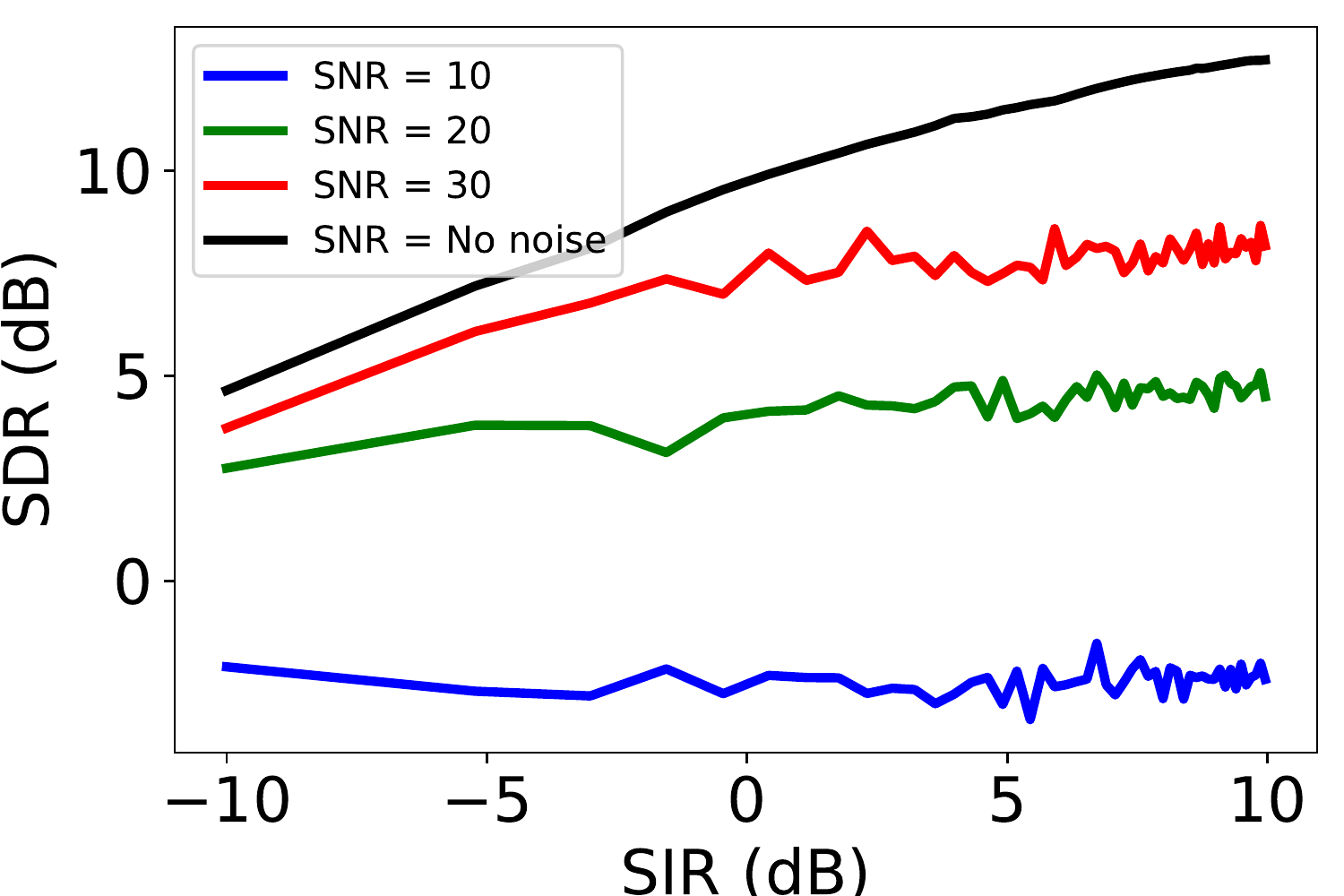}
    }
    \hspace{0.01\textwidth}
    \subfigure[SIR before separation vs SIR after separation.]{
    \label{fig:sir-sir}
    \includegraphics[width=0.43\columnwidth]{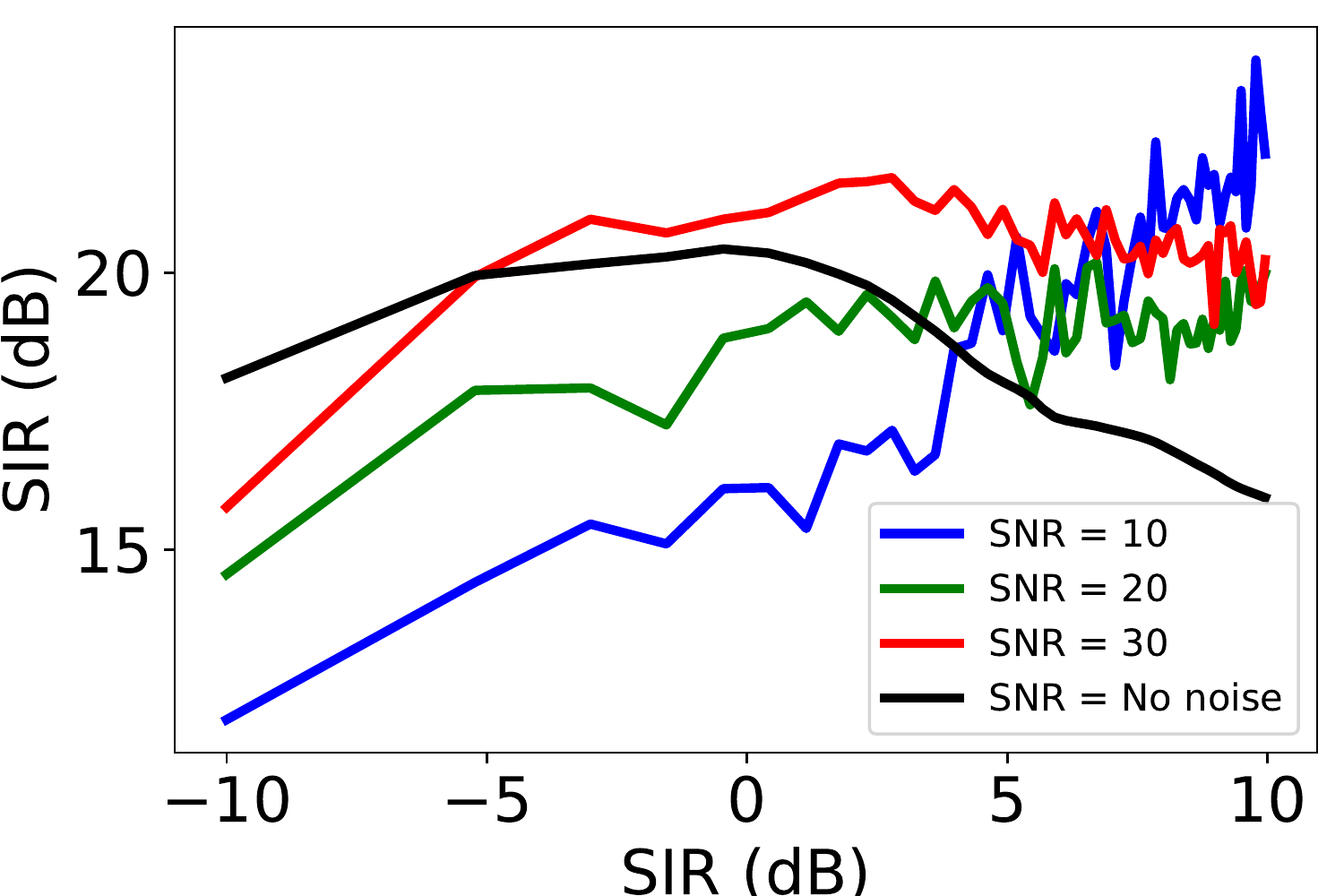}
    }
\hspace{0.001\textwidth}
    \caption{Source separation performance. SDR (a) remains almost constant and SIR (b) is enhanced by source separation.} 
    \label{fig:separation visualization}
\end{figure}


The results are shown in Fig.~\ref{fig:separation visualization}. 
From Fig.~\ref{fig:sir-sdr}, we can see that the SDR after source separation remains almost constant under different SIRs. This simply implies that the source separation algorithm introduces little distortion to the original footsteps, which is notably important for our later identification performance. As a matter of fact, we can barely perceive any distortion when playing the separated footsteps, except some residual voices. 
It is observable that after source separation in Fig.~\ref{fig:sir-sir}, SIR is significantly boosted, indicating a success removal of voice interference. 
\begin{figure}[t]
    \centering

    \subfigure[Time domain waveform.]{
    \label{fig:separation performance in time domain}
    \includegraphics[width=0.43\columnwidth]{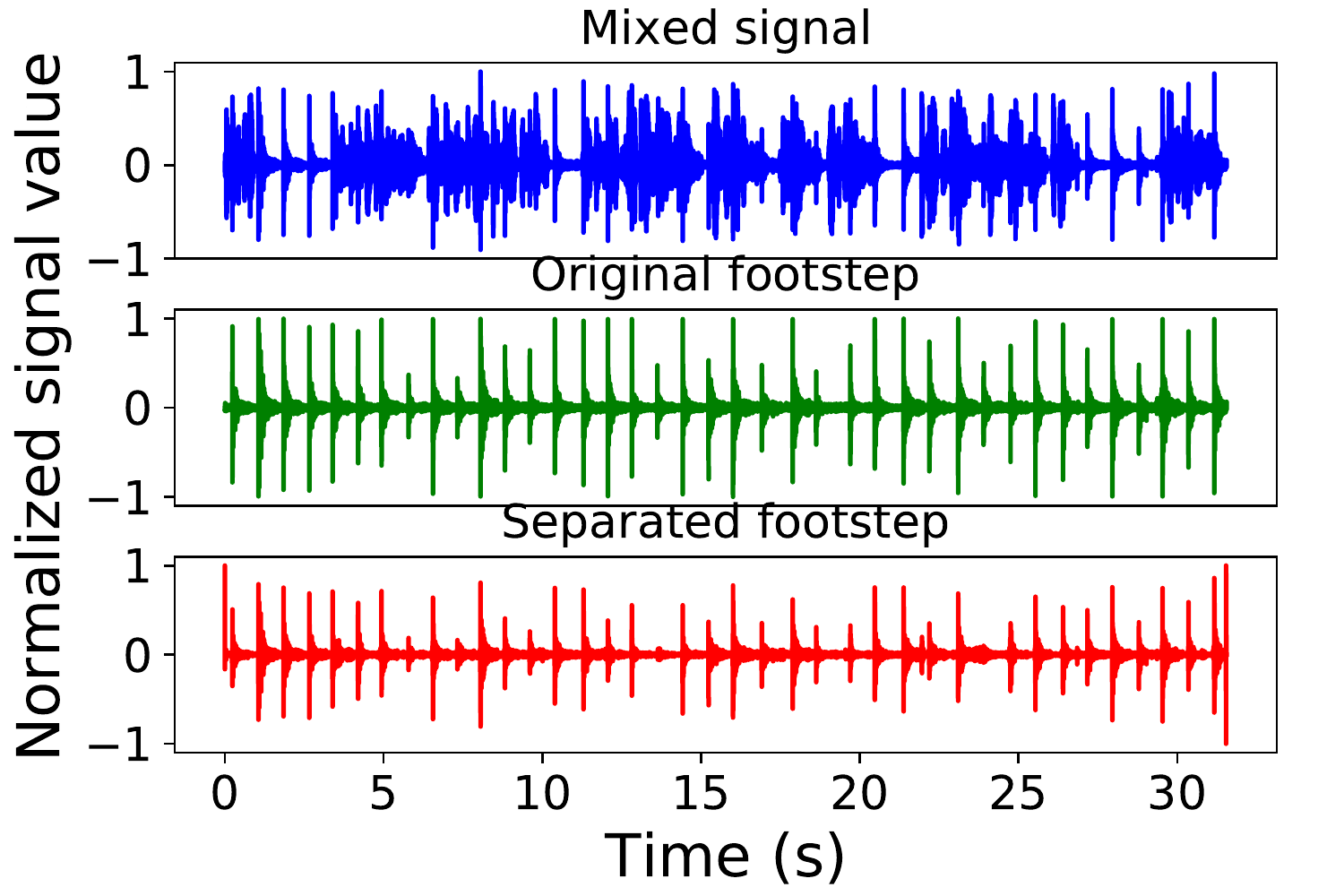}
    }
    \hspace{0.01\textwidth}
    \subfigure[STFT spectrogram.]{
    \label{fig:separation performance in frequency domain}
    \includegraphics[width=0.43\columnwidth]{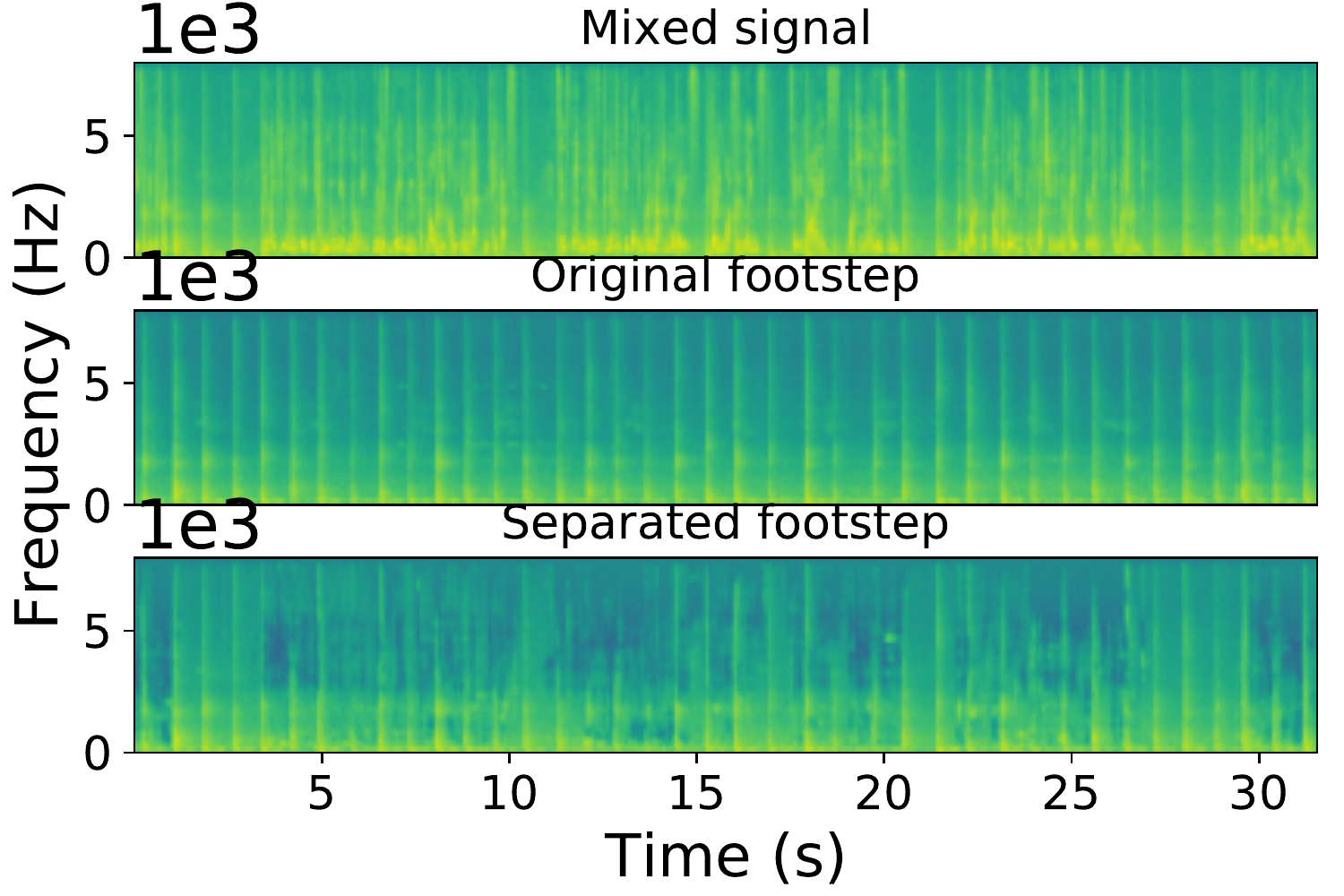}
    }
    \caption{Time domain waveform (a) and STFT spectrogram (b) of mixed, original, and separated footsteps. }
    \label{fig:separation performance}
\end{figure}

We finally showcase waveforms and STFT spectrogram of footsteps after source separation, compared with the mixed and original recorded ones, in Fig.~\ref{fig:separation performance in time domain} and Fig.~\ref{fig:separation performance in frequency domain}, respectively. In this experiments, the maximum voice magnitude is identical to that of footsteps, under which case, a user mostly notice voice but ignores footstep hence interference is strong. Though under severe interference, the separated footstep waveform is already rather clean as we can see from both Fig.~\ref{fig:separation performance in time domain} and Fig.~\ref{fig:separation performance in frequency domain}: minor distortions and residuals may exist, but none of them introduce perceivable artifacts.  

\subsubsection{Denoising Performance}
The denoising network is used to filter out interference from the background subtraction and get rid of residual signals from the source separation. To evaluate the denoising performance, we synthesize mixtures of real-life recorded footsteps and voices under different SNRs and SDRs, generating a total of 1400 sound clips whose duration is within 20~\!s. Then we check the respective SNR and SDR after denosing.  

\begin{figure}[t]
    \centering
    \subfigure[SNR.]{
    \label{fig:snr gain}
    \includegraphics[width=0.43\columnwidth]{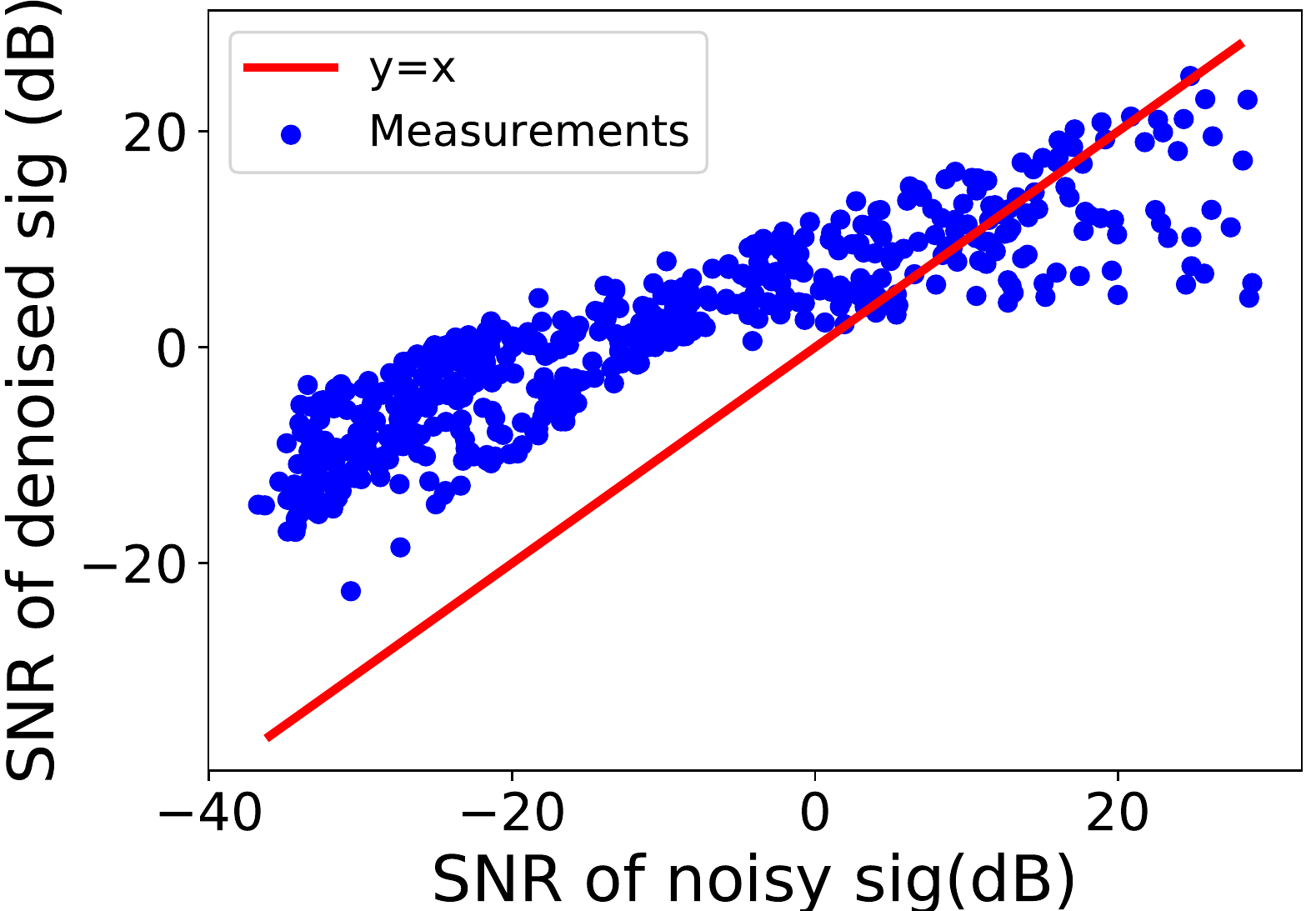}
    }
    \hspace{0.01\textwidth}
    \subfigure[SDR.]{
    \label{fig:sdr gain}
    \includegraphics[width=0.43\columnwidth]{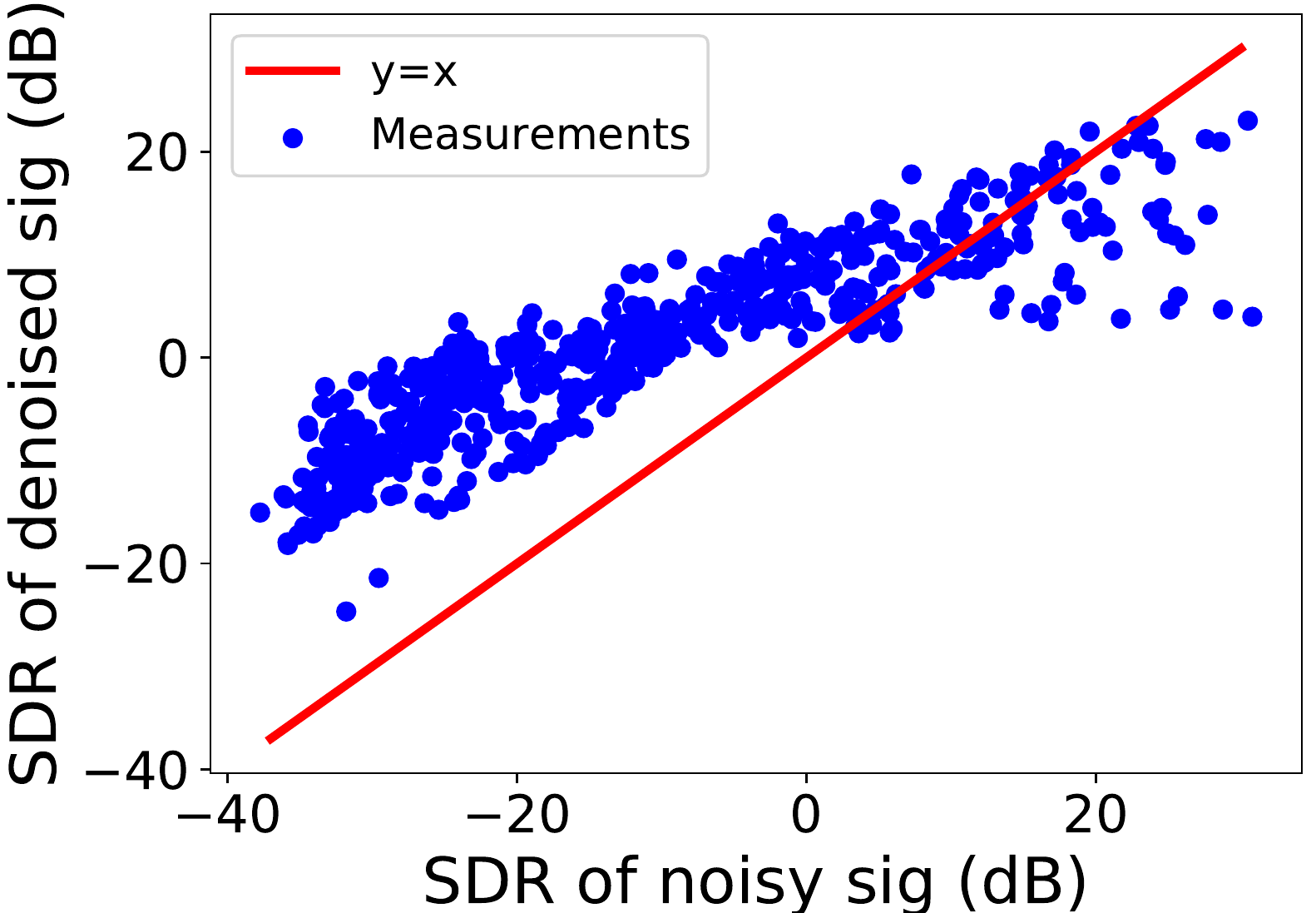}
    }
    \caption{SNR (a) and SDR (b) before and after denoising.}
    \label{fig:snr and sdr after denoising}
\end{figure}

\begin{figure}[b]
    \centering
    \subfigure[Time domain waveform.]{
    \label{fig:denoising at time domain}
    \includegraphics[width=0.43\columnwidth]{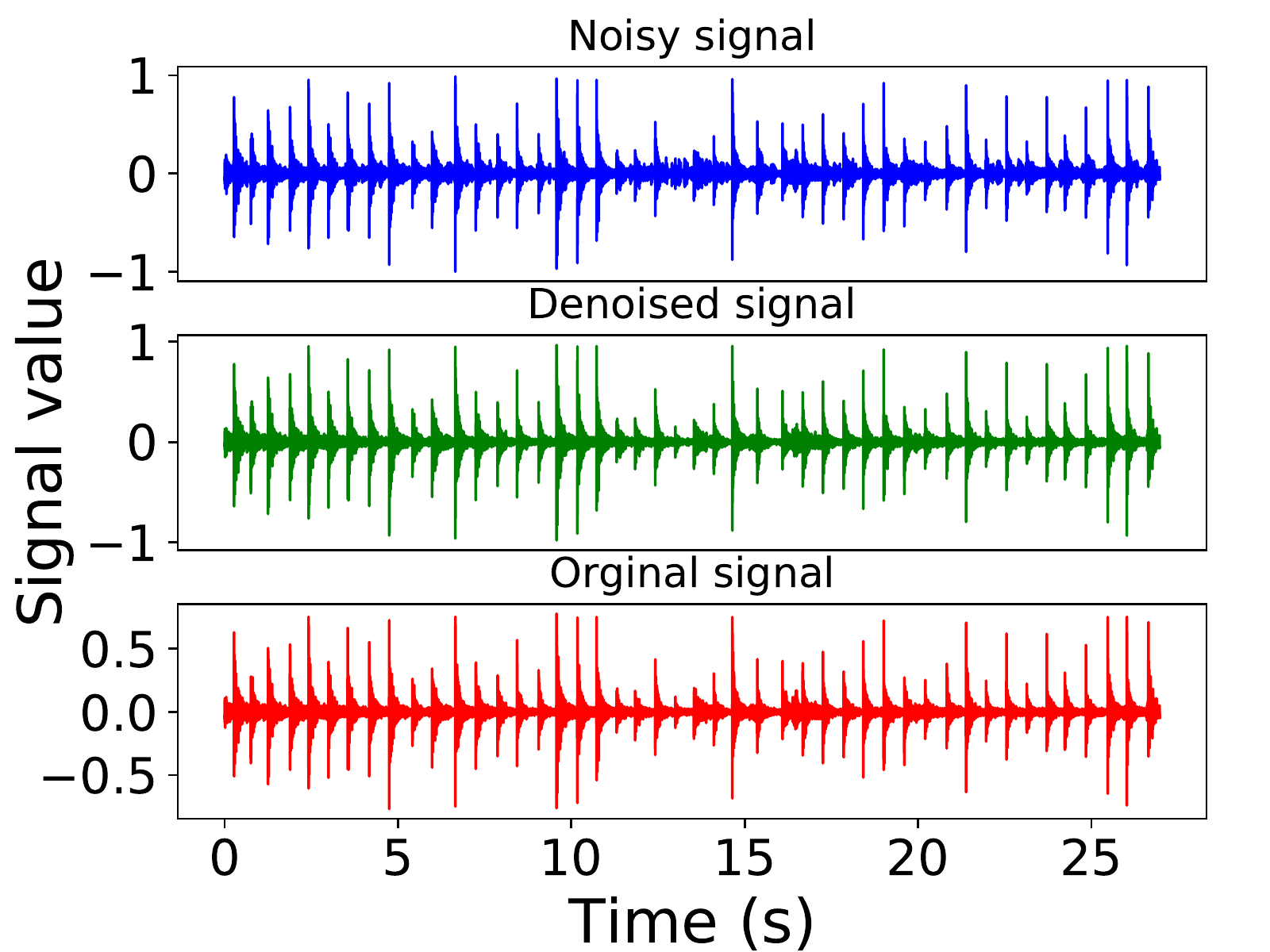}
    }
    \hspace{0.01\textwidth}
    \subfigure[STFT spectrogram.]{
    \label{fig:denoising at frequency domain}
    \includegraphics[width=0.43\columnwidth]{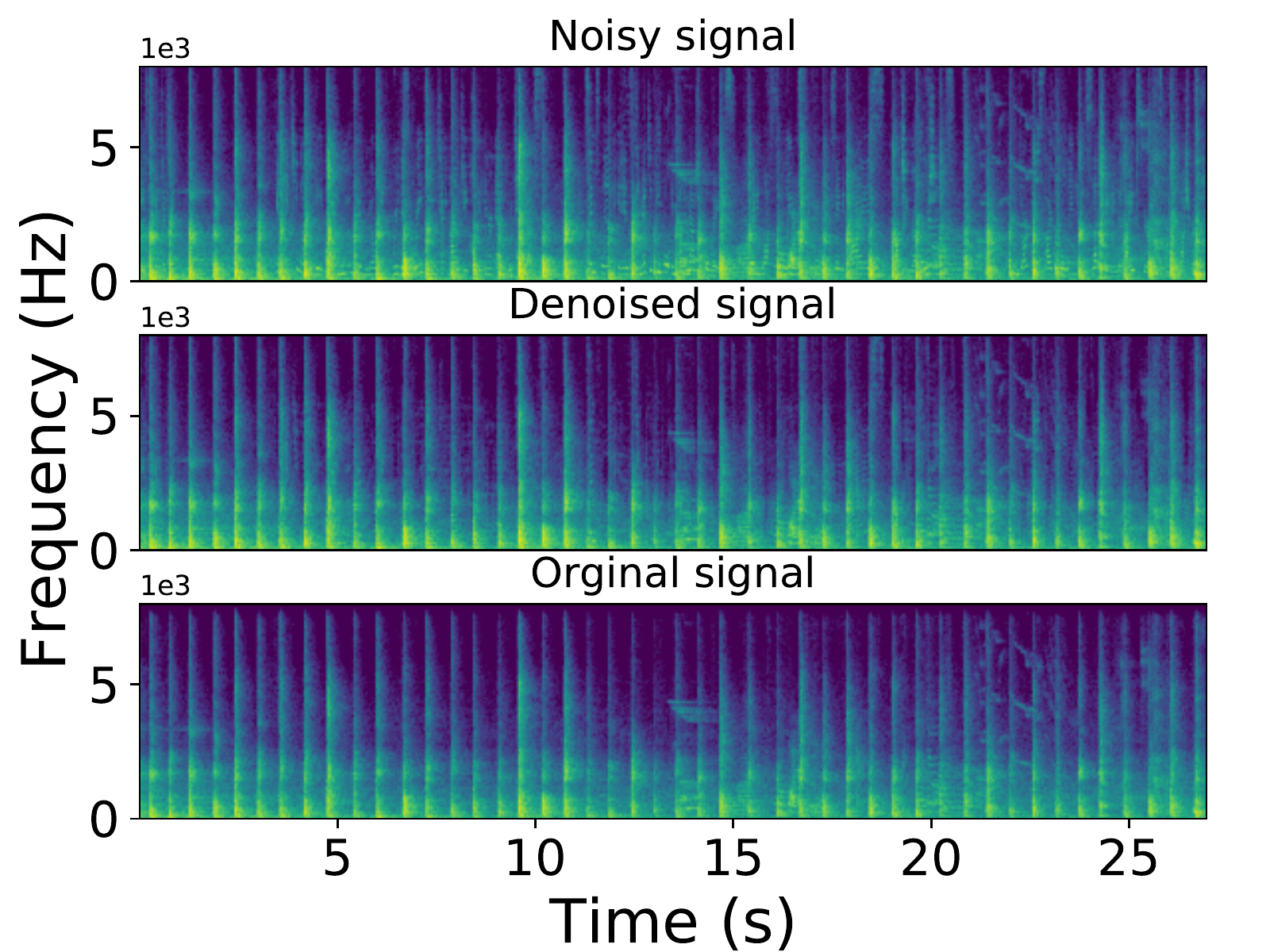}
    }
    \caption{Denoising performance for residual removal. }
    \label{fig:denoising for residual removal}
\end{figure}

Our measurements in Fig.~\ref{fig:snr gain} reveal a maximum SNR gain around 30~\!dB (depending on the background noise type). The SNR gain can be noticeable when the SNR of input noisy signals is relatively low, e.g., below 20~\!dB. However, the network introduces little distortion to its inputs when SNR is high. The same goes
for SDR as shown in Fig.~\ref{fig:sdr gain}. But this little distortion introduces no perceivable difference to the inputs. Meanwhile, we can deactivate the denoising module when SNR or SDR is high so as to prevent the possible distortion since the SNR or SDR can be roughly calculated. We finally showcase our denoising network in residual removal in Fig.~\ref{fig:denoising for residual removal}. 
Both time domain waveform (Fig.~\ref{fig:denoising at time domain}) and Frequency domain spectrogram (Fig.~\ref{fig:denoising at frequency domain}) indicate the success of residual noise removal. The noise removal effect can be visualized by the less signal magnitude variations in time domain waveform and the less ``blurred image'' in spectrogram. 

\subsubsection{Identification Performance}

\begin{figure}[t]
    \centering
    \subfigure[Accuracy over SNR.]{
    \label{fig:accuracy over snr}
    \includegraphics[width=0.43\columnwidth]{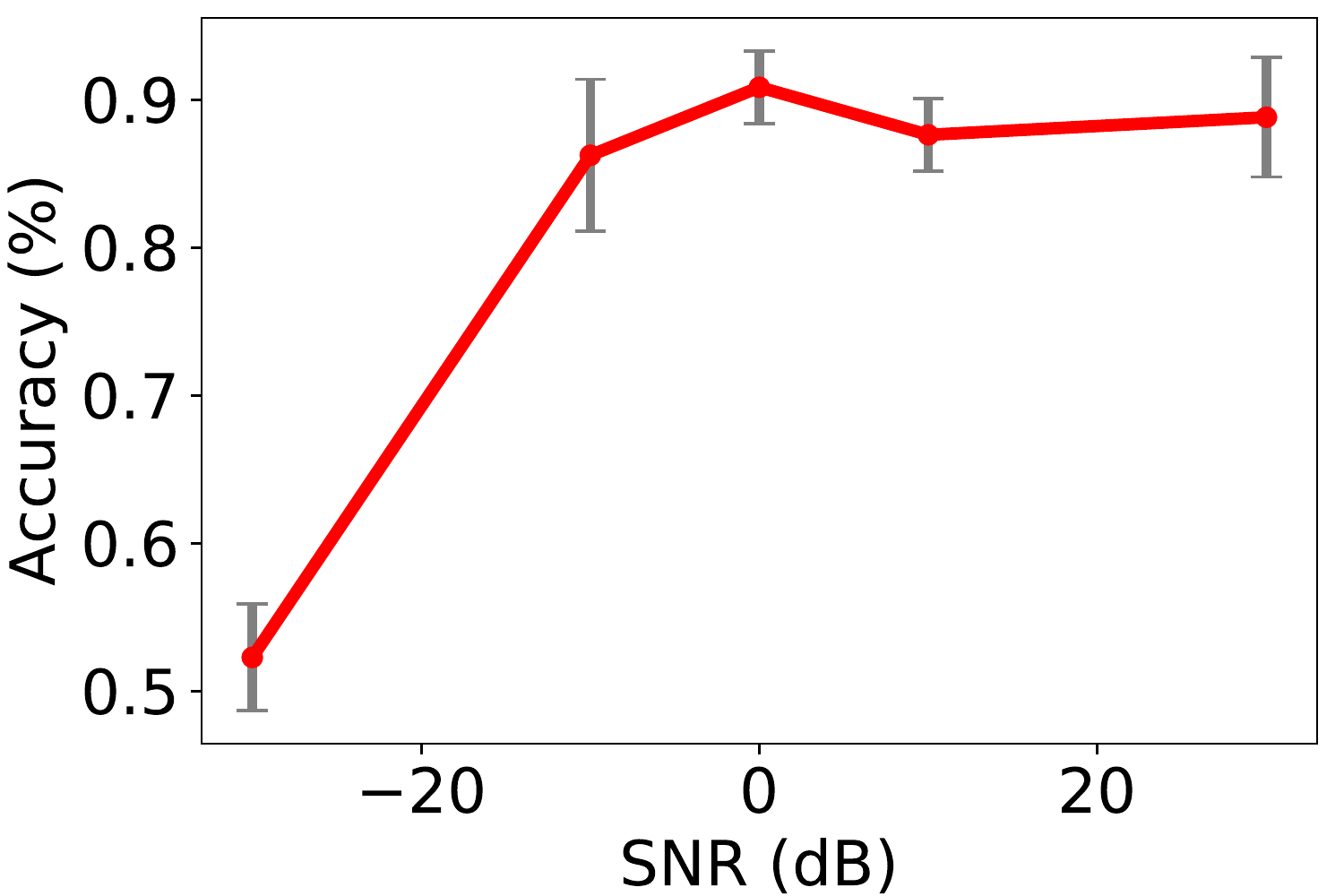}
    }
    \hspace{0.01\textwidth}
    \subfigure[Accuracy over SIR.]{
    \label{fig:accuracy over sir}
    \includegraphics[width=0.43\columnwidth]{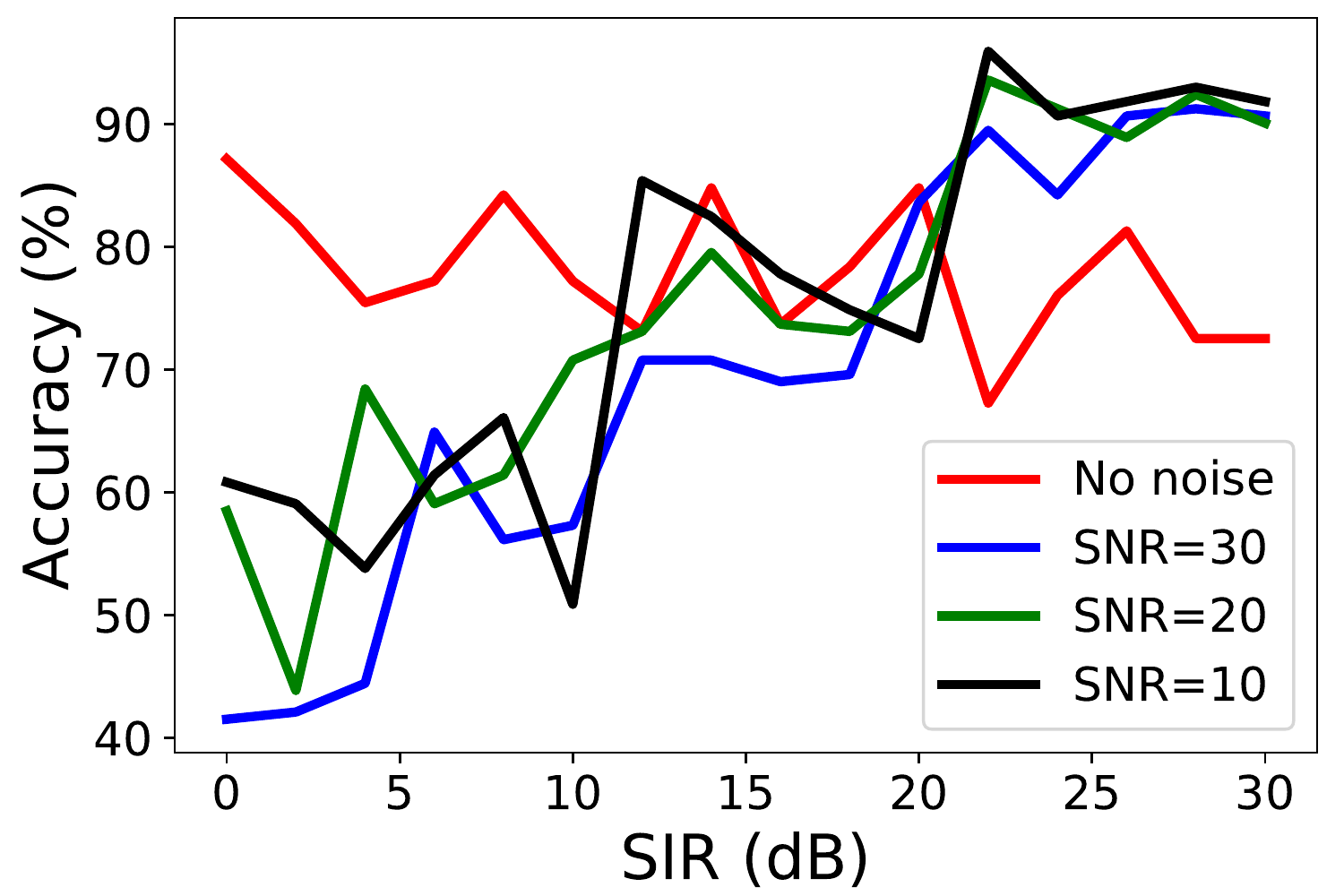}
    }
    \caption{Identification accuracy at various SNRs and SIRs.}
    \label{fig:identification performance}
\end{figure}

We conduct extensive measurements to evaluate the identification performance utilizing real-life recorded footsteps from six users. 
In our first study, we utilize data samples from the same domain (the same environment and walking speed) but only vary the user identity. In this study, we deactive center loss and apply no adversarial learning for ID-Net.  
The results shown in Fig.~\ref{fig:accuracy over snr} reveal that even under a SNR of -12.5~\!dB, 
ID-Net~still achieves 87\% accuracy, demonstrating the feasibility of applying footstep for user identification.
\begin{figure}[b]
    \centering
    \subfigure[Identification accuracy after different processing methods.]{
    \label{fig:run once}
    \includegraphics[width=0.43\columnwidth]{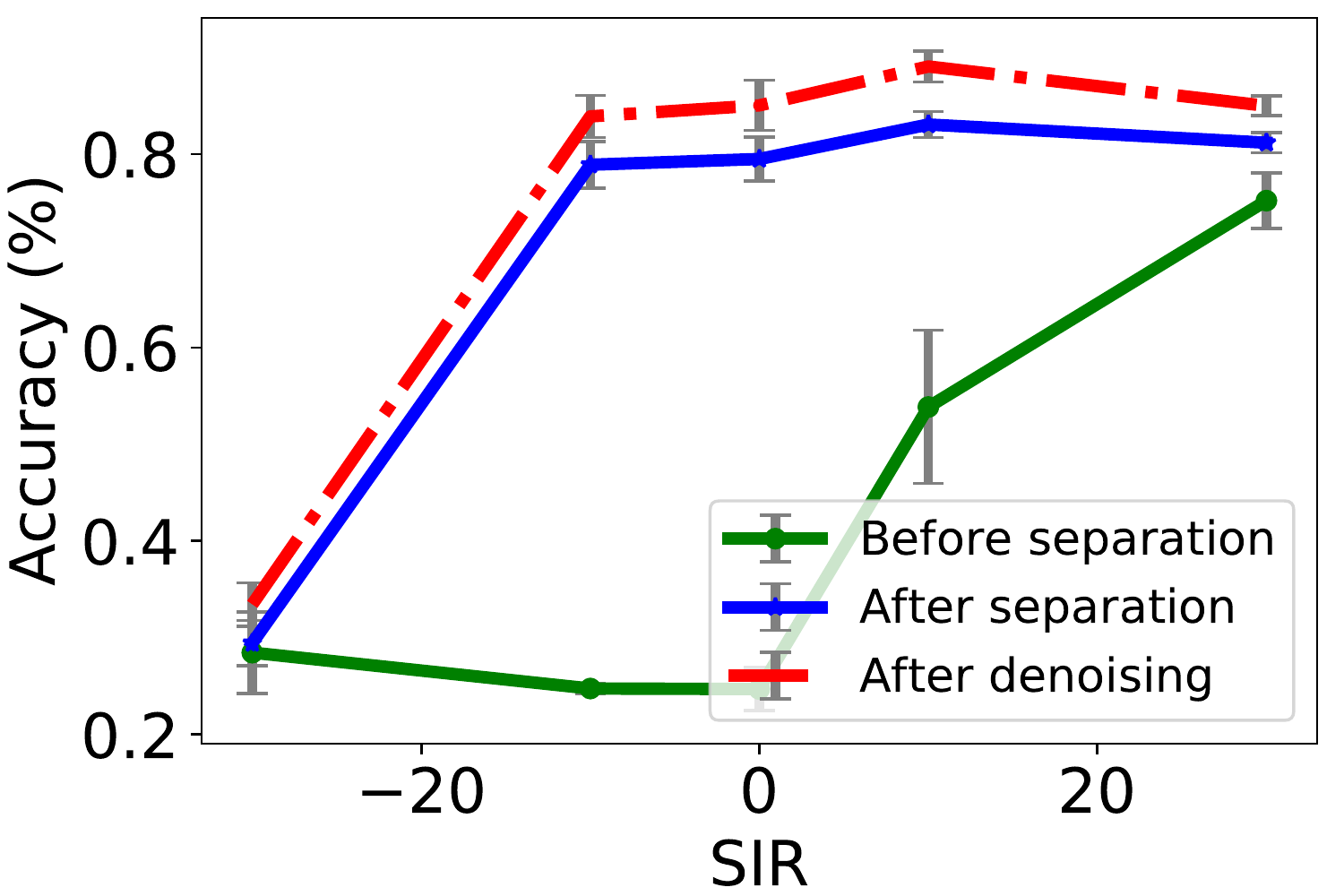}
    }
    \hspace{0.01\textwidth}
    \subfigure[Impact of sampling rate.]{
    \label{fig:impact of sampling rate}
    \includegraphics[width=0.43\columnwidth]{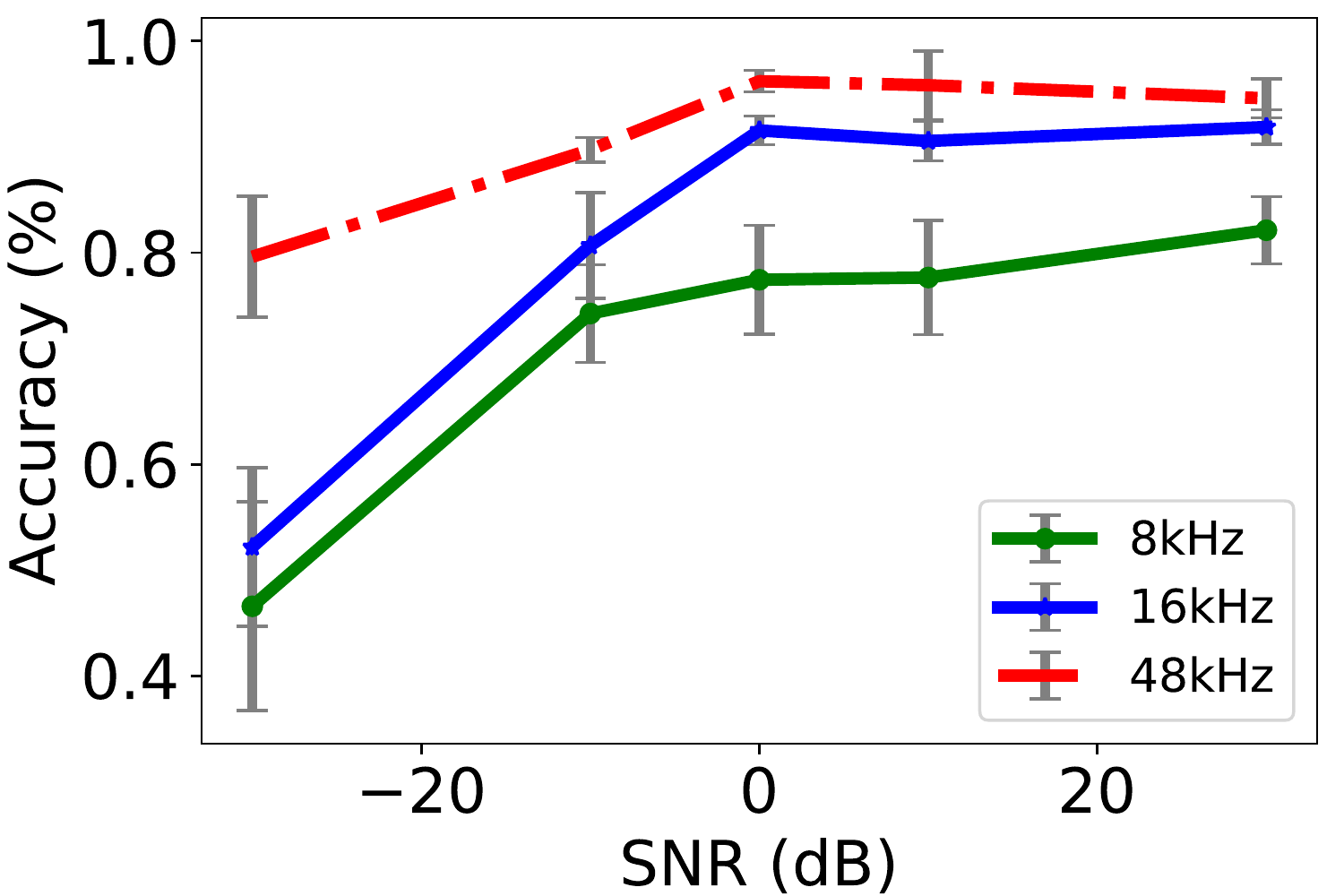}
    }
    \caption{(a) Identification accuracy improved after each processing step. (b) Improving the sampling rate can boost the identification accuracy at low SNR.}
    \label{fig:impacts}
\end{figure}
%
It should be noted that we only utilize one step for identification, if we incorporate multiple footsteps and use a majority voting strategy, the accuracy can be boosted to $1 - C^2_3\times \left(1 - 0.90 \right)^2\times 0.9 - C_3^3(1-0.9)^3 = 97.2$\%.

We then check the identification accuracy under different levels of voice interference (SIR). 
The results in Fig.~\ref{fig:accuracy over sir} show that even under severe interference (SIR = 0), ID-Net can still achieve an accuracy up to 87.13\%. And if adding noise and thereby reducing the SNR for footsteps, the accuracy would drop to 60\%, indicating ID-Net's vulnerability to strong background interference. This also emphasizes the need for source separation and denoising, as the former can deliver SIR gain and the latter provides SNR gain, thereby promoting the identification accuracy. 


We next extensively explore the identification accuracy after source separation and denoising. The results shown in Fig.~\ref{fig:run once} reveals that source separation can achieve a maximum accuracy gain of 59.9\% while denoising network can boost the performance by an average of 5\%. 
We then explore the impact of sampling rate on the final identification performance and Fig.~\ref{fig:impact of sampling rate} shows the results. It can be observed that improving the sampling rate can contribute to better identification performance. The performance gain is marginal when SNR is sufficiently high ($>20$~\!dB), if the sampling rate hits 48~\!kHz. Since a higher sampling rate requires more computational power but achieves little performance improvement, we therefore adopt 16~\!kHz in our system.





\begin{figure}[t]
    \centering
    \subfigure[Low-dimensional features without center loss.]{
    \label{fig:without center loss}
    \includegraphics[width=0.43\columnwidth]{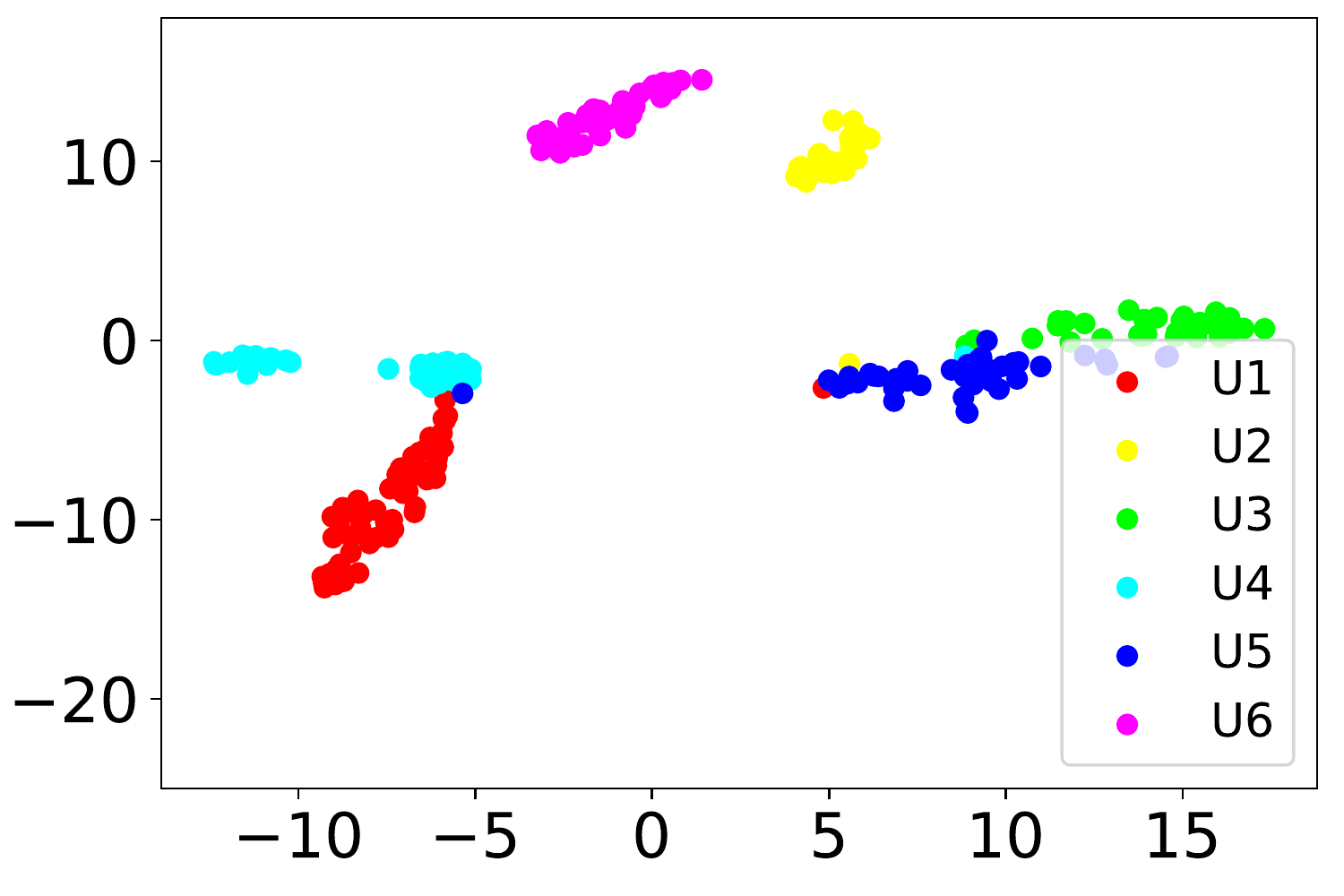}
    }
    \hspace{0.01\textwidth}
    \subfigure[Low-dimensional features with center loss.]{
    \label{fig:with center loss}
    \includegraphics[width=0.43\columnwidth]{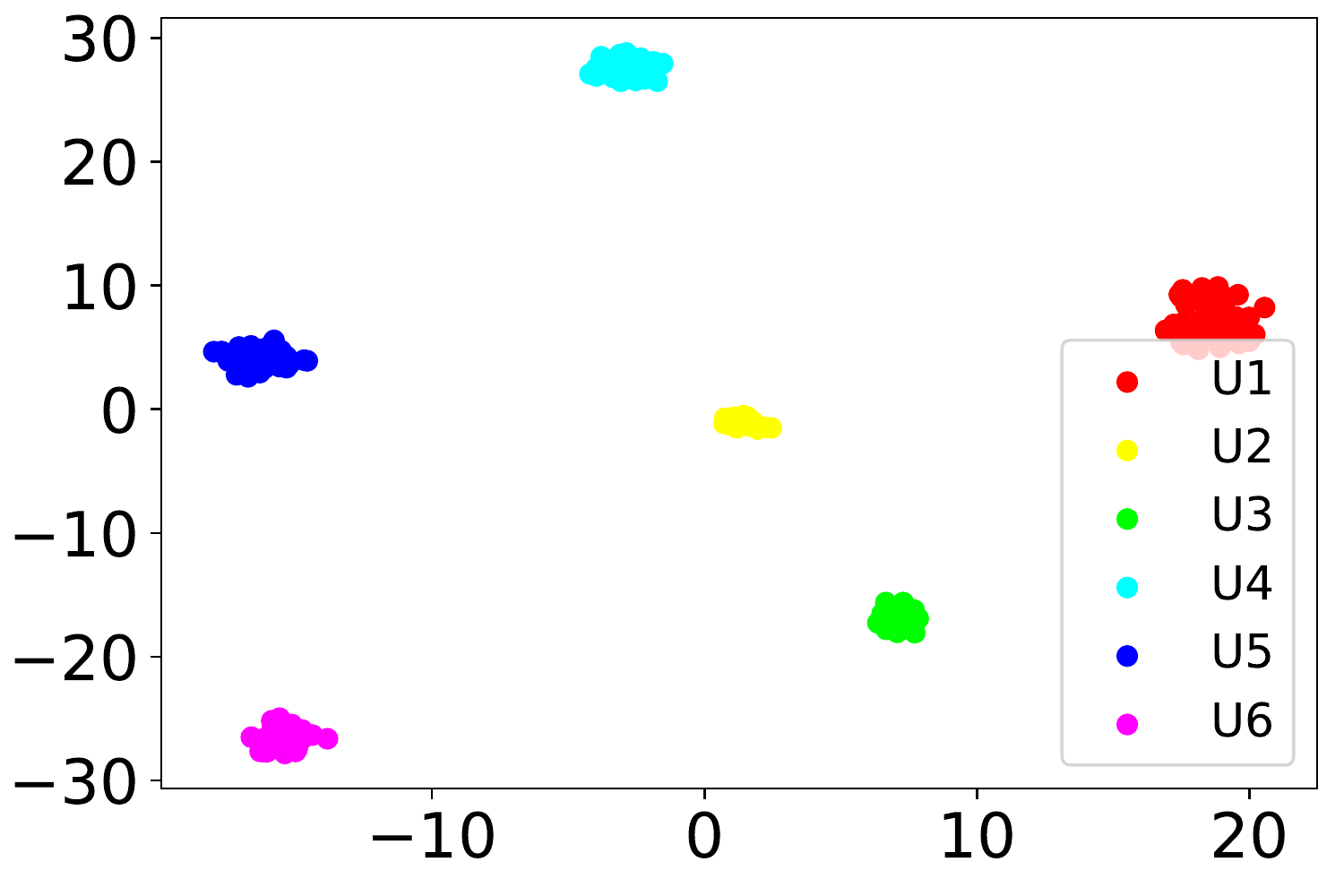}
    }
    \caption{Low-dimensional feature visualization without (a) and with (b) center loss. It clearly shows the power of center loss in  maximizing inter-class boundaries while minimizing intra-class distances.}
    \label{fig:power of center loss}
\end{figure}

We next conduct an ablation study on the effectiveness of center loss, the impact of which on the classified features can be visualized in Fig.~\ref{fig:power of center loss}. It is observable that center loss can effectively maximize the inter-class boundaries and minimize intra-class distances. This ability could not only enhance the identification performance but also improve the generalizability of ID-Net. Applying center loss sometimes can push the identification accuracy to almost 100\%. 

\begin{table}[htp]
\centering
\small
\caption{Accuracy without domain adaptation.}
\label{tab:accuracy under different domains}
\begin{tabular}{|c|c|c|c|}
\hline
\textbf{Accuracy (\%)} & \textbf{Distance} & \textbf{Speed} & \textbf{Environment} \\ \hline
Distance & 76.4 & 57.2 & 9.11 \\ \hline
Speed & 57.2 & 62.01 & 14.31 \\ \hline
Environment & 9.11 & 14.31 & 12.04 \\ \hline
\end{tabular}
\end{table}

We test the identification performance under different domains including speed, distance, and environment variations in the following experiments. 
Specifically, our footsteps are captured under:
1) three levels of walking speed, namely 0.2~\!m/s, 0.5~\!m/s, and 1~\!m/s,
2) different distances ranging from $0$ to $3$~\!m, 
3) heterogeneous environments including common indoor office, home appliance, hall, corridor, etc that exhibit different ground materials and background interference. 

We first show the impact of domains on the identification performance when we deactivate center loss and domain predictor. The results are displayed in Table~\ref{tab:accuracy under different domains} and they tells us that domain conditions can have a notable impact on the identification accuracy. To read the statistics in Table~\ref{tab:accuracy under different domains}, each row and column indicate the number of domains involved in the training data. For instance, $ (\text{row}, \text{column}) = \left( \text{Speed}, \text{Distance} \right) = 57.2\%$ means when the training data involves speed and distance variations, the identification accuracy is 57.2\%. 
According to Eqn.~\eqref{eq:air-borne property}, distance should not impose any negative impacts on the final results. But when data only contains distance variations, the identification accuracy is only 76.42\%. We believe that this is caused by 1) SNR degradation due to propagation loss and 2) structural differences from place to place that cause heterogeneous features. Speed variations, equivalently leading to different impact forces, sabotage the identification accuracy to only 62.01\%. And environment dynamics, introducing different medium properties, undermine the identification accuracy most.

\begin{table}[b]
\centering
\small
\caption{Accuracy with domain adaptation.}
\label{tab:accuracy with domain adaptation}
\begin{tabular}{|c|c|c|c|c|}
\hline
\textbf{Accuracy (\%)} & \textbf{0 Domain} & \textbf{2 D.} & \textbf{2 D.} & \textbf{3 D.}\\ \hline
One footstep & 1 & 88.6 & 84.92 & 81.75\\ \hline
2/3 & 1 & 95.92 & 90.33 & 88.16\\ \hline
3/5 & 1 & 96.53 & 94.47 & 90.73\\ \hline
\end{tabular}
\end{table}

We next explore the identification accuracy under domain adaptation. Particularly, we evaluate the identification accuracy under different number of domains and number of footsteps.  The average results from 100 trials are shown in Table~\ref{tab:accuracy with domain adaptation} and ``2/3'' means we incorporate three steps to identify each user and we accept the result if the same identity appears twice. It can be observable that domain adaptation can significantly improve identification accuracy, pushing the resulting accuracy from a minimal of 9\% (without domain adaptation) to 81.75\%. 
And if incorporating multiple footsteps, the accuracy can be further improved to 90.73\%.

\begin{figure}[t]
    \centering
    \subfigure[Identification accuracy under one footstep interference.]{
    \label{fig:identification accuracy under two footstep overlaped}
    \includegraphics[width=0.435\columnwidth]{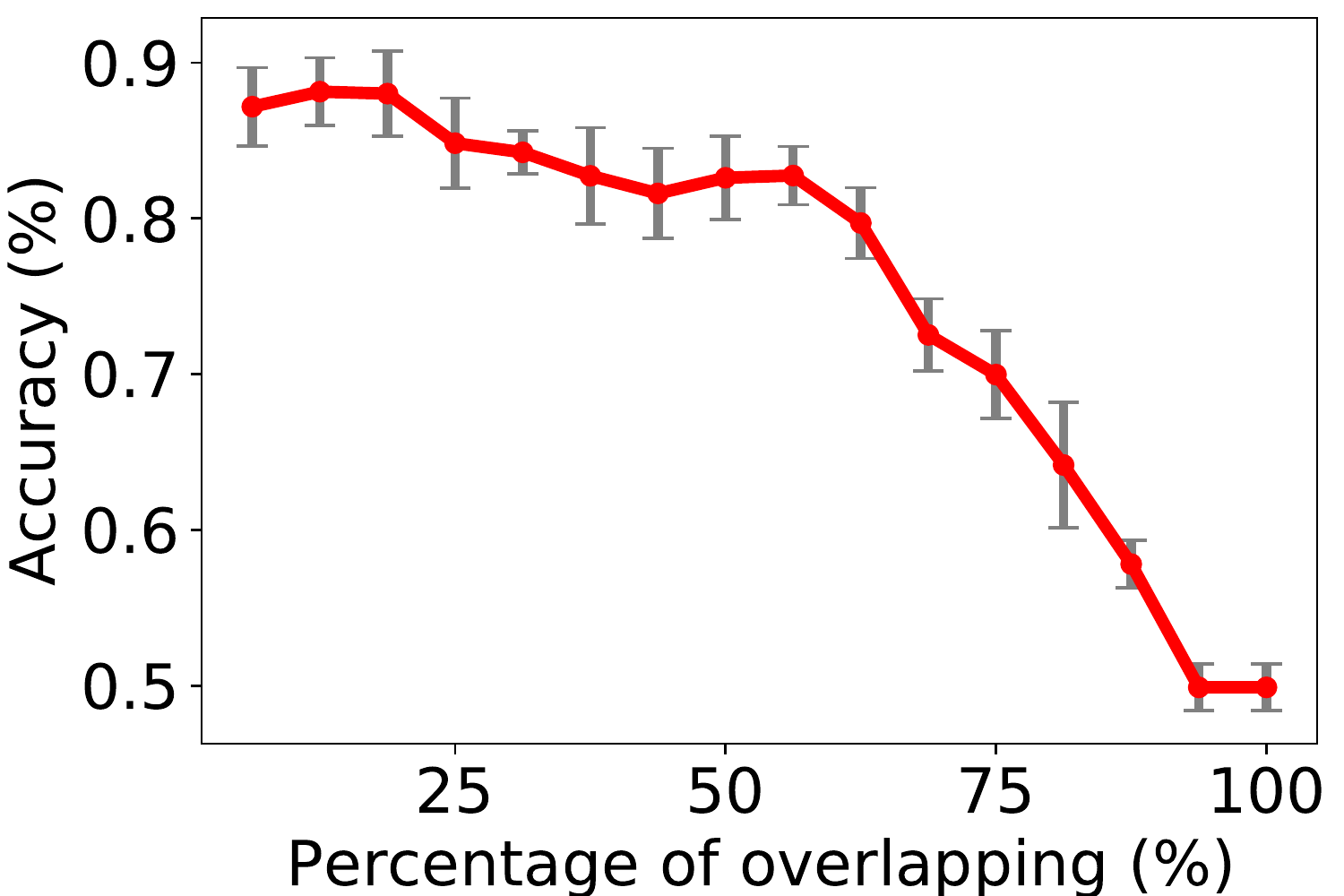}
    }
    \hspace{0.001\textwidth}
    \subfigure[Identification accuracy under multiple footsteps.]{
    \label{fig:identification accuracy under multiple footstep}
    \includegraphics[width=0.435\columnwidth]{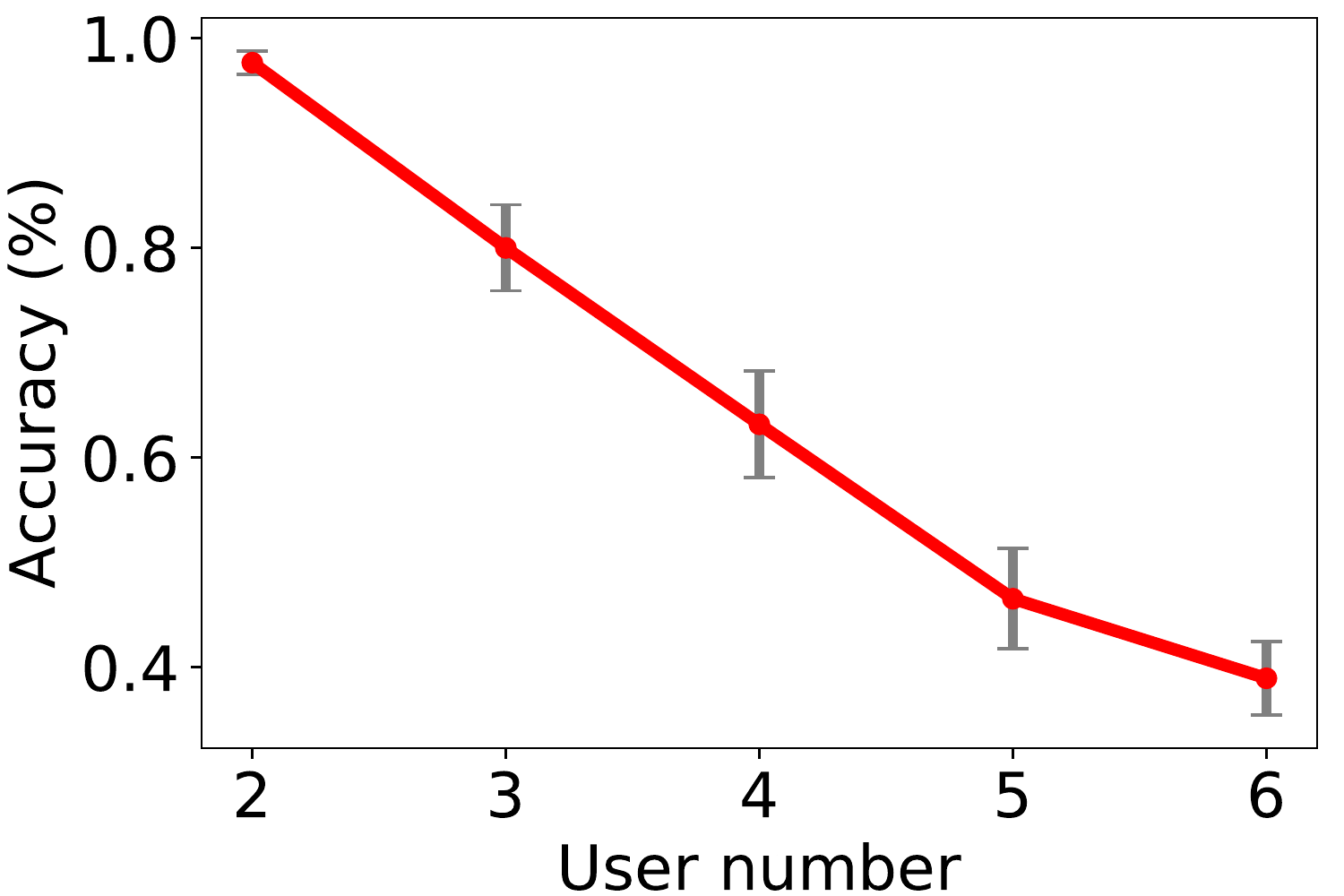}
    }
    \caption{Identification accuracy when one footstep is mixed with partial another footstep components (a). Such a collision can reduce the identification accuracy. Identification accuracy drops when the number of users increases (b). As collision rate increases when more users are involved, the accuracy drops. }
    \label{fig:multiple user identification accuracy}
\end{figure}

We finally explore the identification accuracy under multiple user scenarios. 
We first explore the identification accuracy under the case when each footstep is interfered by only one another footstep. We check the identification accuracy when the footsteps are overlapped at different percentages, the results of which are displayed in Fig.~\ref{fig:identification accuracy under two footstep overlaped}. As the figure tells, the identification accuracy drops monotonically if the percentage of overlapped region increases. However, the accuracy is still around 50\% if two steps are totally overlapped. This simply implies that ID-Net can still recognize these two footsteps but is unable to distinguish them. We then evaluate the performance when multiple person randomly walk in an indoor meeting room where we place the microphone array in the center. The results in Fig.~\ref{fig:identification accuracy under multiple footstep} show that an identification accuracy around 80\% can still be achieved even when there are three users. As the number of user increases, the  collision between different footsteps happens more frequently hence worse performance.

\subsubsection{Defend Against Replay Attack}
\systemname~leverages R-Net to extract spatial information, including range and AoA, from multiple consecutive footsteps to defend against replay attack.  In this section, we first present the performance of R-Net in spatial information extraction and then inspect the defending performance based on these signatures. We run over 1000 trials in an indoor office $6.8\times4.2$~\!m$^2$ where we place the microphone array in the center.

\begin{figure}[t]
    \centering
    \subfigure[Ranging performance.]{
    \label{fig:ranging performance}
    \includegraphics[width=0.44\columnwidth]{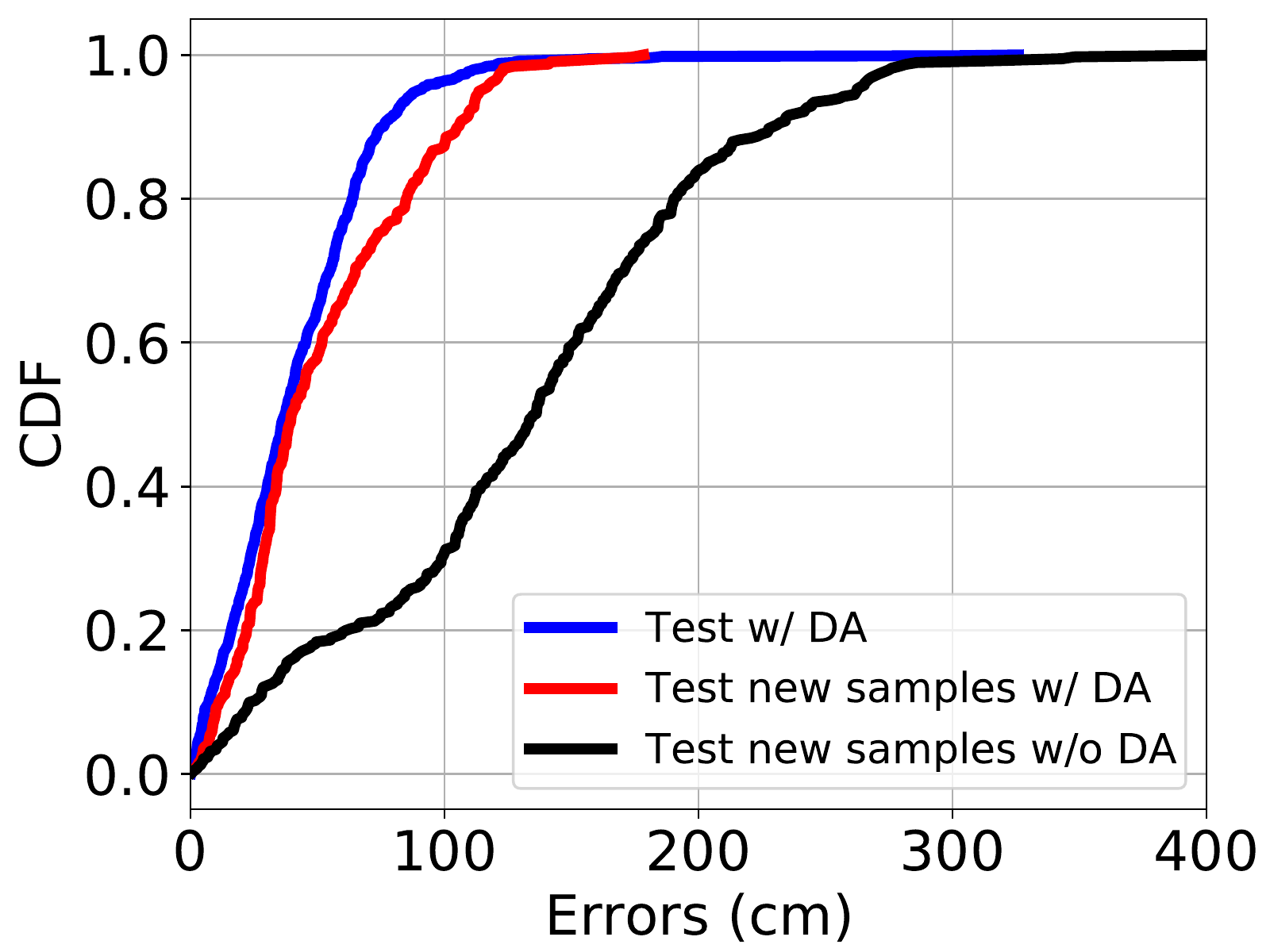}
    }
    \hspace{0.001\textwidth}
    \subfigure[AoA estimation performance.]{
    \label{fig:AoA performance}
    \includegraphics[width=0.45\columnwidth]{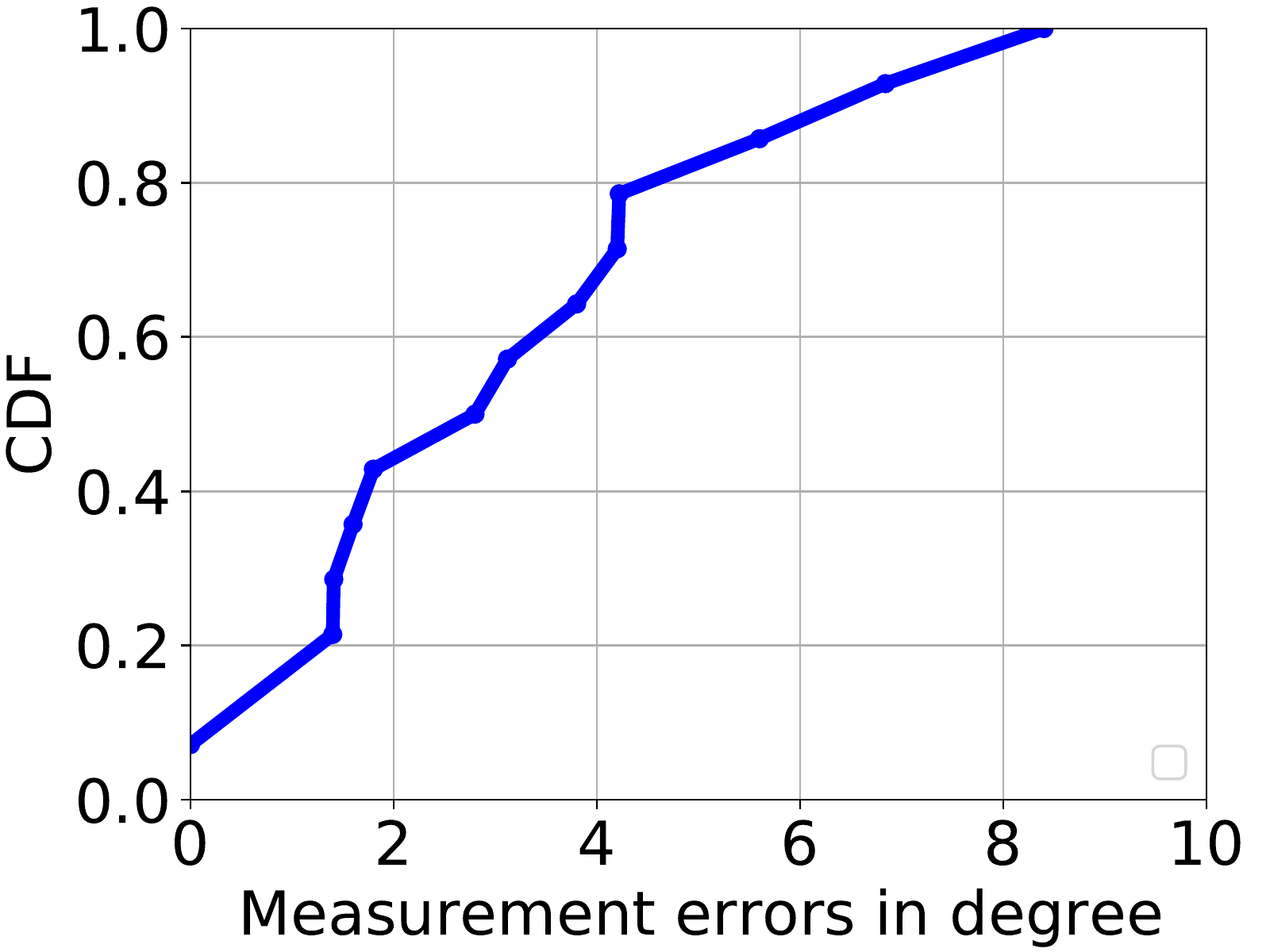}
    }
    \caption{Performance in spatial clues extraction.  }
    \label{fig:spatial clues}
\end{figure}

Fig.~\ref{fig:spatial clues} shows the ranging and AoA estimation performance. In Fig.~\ref{fig:ranging performance}, We verify the ranging performance of R-Net under three cases to demonstrate its capability in domain adaptation. First, we utilize training data (70\% of all data) from all the identities and domains to train R-Net, and we then verify the ranging performance using test data (the remaining 30\%), which we refer to as Test with Domain Adaptation (Test w/DA). Second, we randomly remove one identity from the training data and after training, we apply inference on this particular identity, referred as Test new samples with Domain Adaptation (Test new samples w/DA). Third, we cut the domain predictor from R-Net and apply inference using the same setting with the second case, denoted as Test new samples without Domain adaptation (Test new samples w/o DA). The results in Fig.~\ref{fig:ranging performance} exhibit an median error of around 0.3~\!m, even testing on samples that come from other domains and never participant in the training process. And if without domain adaptation, the median ranging errors would reach up to 1~\!m. The comparison of afore-mentioned results clearly demonstrates the salient performance of R-Net in domain adaptation, as well as in ranging. Fig.~\ref{fig:AoA performance} shows AoA estimation errors are below $10^\circ$. These salient performance lays the foundation of our defending mechanism against replay attack. 

We have checked our defending mechanism under several replay attack scenarios including different walking trajectories and hacking positions. Our measurements reveal that if the trajectories contain complex shapes such as ``L'' or circle shape segments, the detected $\pi$ could easily violates the threshold $\bar{\pi}$ so that we achieve 100\% success in defending these attacks. When there involves only straight line trajectories, 
the variance of detected AoA revealed by replayed attacks never exceeds the preset $10^\circ$ threshold while live footsteps reveal a minimum variation around 32.8$^\circ$ due to location swing caused by the alternation between left and right legs. 
In conclusion, \systemname\ can successfully defend against replay attack.   

%

\section{Related Work}
\label{sec:related work}
In this section, we survey the literature on user identification. 
Whereas common identification techniques have a broad categories, ranging from traditional computer vision, fingerprint sensing, and iris scan, they are rather irrelevant to our proposal. Therefore, we shall not review these common techniques but focus only on solutions that leverage emergent sensing techniques and adopt behavioral biometrics for user identification. 

The proposals of~\cite{WiFiID,WiWho,WiFiSensing1,WiFiSensing2} exploit the gait information to identify users during their walking. The basic idea behind these systems is that the particular walking cycle of each user can be sampled by WiFi signals. However, they may not be able to adapt to environment dynamics, and even walking direction variations can severely affect the identification accuracy. Meanwhile, they often fail to work in practice due to the severe interference from WiFi’s main function of communications and other co-spectrum devices.

FootprintID~\cite{FootprintID} is a structural vibration based identification system. It employs Gephone~\cite{Gephone} to sense the structural vibration caused by a footstep. For identification, it again relies on the gait patterns extracted from multiple structural vibration measurements; the reported identification accuracy for 10 people may reach up to 96\%. However, this promising solution still leaves many open issues, including sensor location variation, multiple pedestrians interference, footwear variation. Other similar behavioral biometrics enabled identification system can be found in~\cite{GaitAccelerometer1,GaitAccelerometer2,GaitAccelerometer3}, a well as~\cite{Footstep3} where accelerometer and camera are used together as sensors. 

To summarize, existing technologies driven by behavioral biometrics often require multiple measurements hence long latency identification experience. On the contrary, \systemname\ solves this problem elegantly by requiring as few as only one step. While \systemname\ can be deemed as a type of behavioral biometrics, it is actually quite related to voiceprint recognition~\cite{VoiceRecognition,VoiceRecognition1}; it can be deemed as a ``footstep-print'' enabled identification system. \systemname\ is similar to those acoustic fingerprint based systems~\cite{VoiceRecognition,VoiceRecognition1} but \systemname~is totally passive and thus can provide better user experience. 
The most similar work to \systemname\ is the one from~\cite{Footstep1} where footstep patterns rather than gaits are used for identification. But this proposal requires an excessive number of piezoelectric sensors to capture footstep signals while \systemname\ utilizes only commodity microphone, significantly reducing the deployment cost and rendering itself widely applicable for indoor scenarios.   

\section{Conclusion}
\label{sec:conclusion}
In this paper, we have explored the possibility of exploiting footsteps for passive user identification. We have proposed \systemname\ as a multi-person identification system driven by a pipeline of signal processing and deep learning techniques. \systemname\ demands as few as a single footstep to enable user identification and is immune to replay attacks. \systemname\ is even feasible to work under continuous voice interference, thanks to a novel source separation and denoising network. To have \systemname\ working across different domains, we have exploited domain adversarial adaptation scheme with a center loss to further enhance its generalization ability across different domains. We have implemented a prototype for \systemname\ and extensively evaluated its performance; the results confirm that \systemname\ achieves a cross-domain identification accuracy up to 90\%. Since \systemname\ outperforms existing passive identification system in both deployment cost and identification latency, we have the reason to believe that \systemname\ has the potential for a wide adoption.


\bibliographystyle{IEEEtran}
\bibliography{mybib}

\end{document}